%% file: main.tex
\documentclass[10pt,twocolumn,letterpaper]{article}
\usepackage[pagenumbers]{jopano}
\definecolor{myblue}{rgb}{0.22,0.51,0.75}
\usepackage[pagebackref,breaklinks,colorlinks,allcolors=myblue]{hyperref}

\usepackage{multirow}
\usepackage{placeins}
\usepackage{stfloats}

\usepackage{utils}
\usepackage{tabularx}
\usepackage[misc,geometry]{ifsym}
\NewDocumentCommand{\chzl}{o m}{\comm[HZL Comment][red][#1]{#2}}
\NewDocumentCommand{\ccyw}{o m}{\comm[CYW Comment][magenta][#1]{#2}}
\NewDocumentCommand{\addcyw}{m o}{\add[violet][CYW Comment]{#1}[#2]}
\NewDocumentCommand{\strkcyw}{m o}{\strk[violet]CYW Comment]{#1}[#2]}
\NewDocumentCommand{\delcyw}{m o}{\del[violet][CYW Comment]{#1}[#2]}
\NewDocumentCommand{\rplcyw}{m m o}{\rpl[violet][CYW Comment]{#1}{#2}[#3]}

\title{OSI: One-step Inversion Excels in Extracting Diffusion Watermarks\vspace{-1ex}}

\author{
Yuwei Chen\textsuperscript{1,2} \quad Zhenliang He\textsuperscript{1,2\>\Letter} \quad Jia Tang\textsuperscript{1,3} \quad Meina Kan\textsuperscript{1,2} \quad Shiguang Shan\textsuperscript{1,2}\vspace{1ex}\\
\textsuperscript{1}State Key Laboratory of AI Safety, Institute of Computing Technology, CAS, China\\
\textsuperscript{2}University of Chinese Academy of Sciences (CAS), China\\
\textsuperscript{3}School of Information Science and Technology, ShanghaiTech University, China\vspace{0.25ex}\\
\tt\small \{yuwei.chen,jia.tang\}@vipl.ict.ac.cn \quad \{hezhenliang,kanmeina,sgshan\}@ict.ac.cn\vspace{-1.25ex}
}

\begin{document}
\maketitle

\begingroup
\renewcommand\thefootnote{\fnsymbol{footnote}}%
\footnotetext{\Letter~Corresponding author.}%
\endgroup

\input{sec/0_abstract}
\input{sec/main_text}

{
    \small
    \bibliographystyle{ieeenat_fullname}
    \bibliography{main}
}

\appendix
\input{sec/X_suppl}
\end{document}

%% file: sec/0_abstract.tex
\begin{abstract}
Watermarking is an important mechanism for provenance and copyright protection of diffusion-generated images. 
Training-free methods, exemplified by Gaussian Shading, embed watermarks into the initial noise of diffusion models with negligible impact on the quality of generated images. 
However, extracting this type of watermark typically requires \textbf{multi-step diffusion inversion} to obtain precise initial noise, which is computationally expensive and time-consuming. 
To address this issue, we propose \textbf{One-step Inversion (OSI)}, a significantly faster and more accurate method for extracting Gaussian Shading style watermarks. 
OSI reformulates watermark extraction as a learnable sign classification problem, which eliminates the need for precise regression of the initial noise. 
Then, we initialize the OSI model from the diffusion backbone and finetune it on synthesized noise-image pairs with a sign classification objective. 
In this manner, the OSI model is able to accomplish the watermark extraction efficiently in only one step. 
Our OSI substantially outperforms the multi-step diffusion inversion method: it is $20\times$ faster, achieves higher extraction accuracy, and doubles the watermark payload capacity. 
Extensive experiments across diverse schedulers, diffusion backbones, and cryptographic schemes consistently show improvements, demonstrating the generality of our OSI framework.

\end{abstract}

%% file: sec/main_text.tex
\vspace{-1.85ex}
\section{Introduction}
\vspace{-0.85ex}
\label{sec:intro}
The rapid advancement of diffusion models has catalyzed the emergence of numerous high-quality image generation platforms~\cite{runway_website, midjourney_website, klingai_website}, significantly enhancing creative capabilities in both professional and consumer domains.
However, the risks related to intellectual property, authorship, and accountability have also been increasing concurrently. 
Therefore, it is imperative to develop technical mechanisms to safeguard the copyrights of synthesized images and trace their unauthorized usage.

Over the past years, digital watermarking has proven to be an effective mechanism for image protection and provenance~\cite{adigitalwatermark_1994,dct_1998,qim_2001}. 
In the generative era, watermarking remains a promising countermeasure for safeguarding diffusion-generated images, especially because the watermarking mechanism can be deeply integrated into the generation process.
Training-free methods~\cite{treering_2023,gaussianshading_2024} have emerged as one of the most promising paradigms for diffusion watermarking.
These methods embed watermarks into the initial noise of diffusion models, which has a negligible impact on the quality of the generated images since neither the model parameters nor the generation procedure are modified.
As a result, training-free approaches have become an active research focus and have undergone numerous refinements~\cite{zodiac_2024,ringid_2024,prcwatermark_2025,gaussmarker_2025,tagwm_2025,maxsive_2025,secureandefficient_2025,sfwmark_2025,t2smark_2025, gaussianshading++_2025,swaldm_2025}.

Despite the promise of training-free watermarking, representative methods such as Gaussian Shading~\cite{gaussianshading_2024} rely on \textit{multi-step diffusion inversion} to extract watermarks, which incurs substantial computational overhead, particularly for large-scale deployment.
A crucial asymmetry has been overlooked: while multi-step diffusion inversion can recover the \textit{exact value} of the initial noise, the watermark extraction task is only concerned with the \textit{discrete signs} of the noise.
Therefore, we argue that adopting multi-step diffusion inversion for extracting Gaussian Shading style watermarks is ``using a sledgehammer to crack a nut'', which introduces unnecessary computational complexity and latency. 
In other words, we believe that there exists a simpler and more efficient solution for watermark extraction.
Notably, recent studies have developed one-step diffusion generation methods~\cite{onestepdiffusion_2024, onestepshortcut_2025,meanflow_2025}, which significantly reduce the computational cost of the image generation process.
Conversely, \textit{is it possible to develop a one-step inversion approach specialized for efficient watermark extraction?}

Based on the above analysis, we suggest that the extraction of Gaussian Shading style watermarks is more appropriately formulated as a \textit{discrete classification} problem rather than a \textit{continuous regression} problem. 
Accordingly, we propose \textbf{One-step Inversion (OSI)} framework, which extracts a watermark through a single forward pass.
Specifically, we first initialize the OSI model from the diffusion backbone to leverage its inherent capability of exact noise inversion~\cite{ddim_2021,gaussianshading_2024,prcwatermark_2025}.
Next, we construct a supervised dataset by using a diffusion pipeline to synthesize noise-image pairs.
Finally, we finetune the OSI model on these data pairs with a sign classification objective, resulting in a one-step model for watermark extraction.
In this manner, we achieve significant improvements in watermark extraction compared with previous multi-step diffusion inversion method, specifically: 1) a $20\times$ increase in extraction speed, 2) improved extraction accuracy, and 3) a double increase in watermark payload capacity.
Furthermore, extensive experiments across diverse schedulers, diffusion backbones, and cryptographic schemes consistently show improvements, demonstrating the generality of our OSI framework.
These properties make OSI a practical and promising approach in real-world deployment.

\section{Related Works}
\subsection{Finetuning-based Diffusion Watermarks}
Finetuning-based diffusion watermarking methods are intrinsically tailored to the diffusion architecture.
A representative group of approaches update parameters within the original generation pipeline: WatermarkDM~\cite{watermarkdm_2023} trains the diffusion model on specially watermarked image datasets, and Stable Signature~\cite{stablesignature_2023} finetunes the decoder to embed watermarks during latent decoding. 
Other approaches integrate auxiliary learnable modules: AquaLoRA~\cite{aqualora_2024} attaches learnable LoRA~\cite{lora_2022} to the diffusion UNet to embed watermark, and ROBIN~\cite{robin_2024} learns additional watermark parameters and learnable conditioning signals to modify intermediate sampling steps.
Since these finetuning-based methods intervene in the generation process, they require careful balancing of training objectives, schedules, and parameterization to avoid severely degrading image quality.

\subsection{Training-Free Diffusion Watermarks}
Training-Free methods embed a watermark into the initial noise, thereby avoiding any modification to the diffusion model.
These approaches can be broadly grouped into two families: frequency-domain watermark (Tree-Ring style) and spatial-domain watermark (Gaussian Shading style). 
\paragraph{Tree-Ring Style watermarks}
Tree-Ring~\cite{treering_2023} embeds watermarks by \textit{modifying the frequency statistics of the initial latent}, imposing concentric ring patterns that can be extracted via multi-step diffusion inversion.
Although the generator remains training-free, such frequency-domain perturbations deviate from the Gaussian prior and may reduce image fidelity.
Subsequent works~\cite{zodiac_2024, ringid_2024, sfwmark_2025} refine the frequency patterns to better align with the prior distribution, aiming to reduce the perceptual impact while preserving watermark robustness. 

\paragraph{Gaussian Shading Style watermarks}
Gaussian Shading (GS)~\cite{gaussianshading_2024} introduces a widely adopted spatial-domain paradigm. It bridges diffusion and cryptography by applying \textit{sign masks} derived from the watermarks to control the signs of the initial latent, which is equivalent to \textit{assigning a fixed subset of the latent space to each watermark}. 
GS employs repetition codes (REP) to enhance the robustness of watermark. 
PRCW~\cite{prcwatermark_2025} extends GS framework with pseudo-random error-correcting codes (PRC)~\cite{prcode_2024} to enhance undetectability. 
A series of follow-up studies ~\cite{gaussianshading++_2025,gaussmarker_2025,maxsive_2025,secureandefficient_2025,swaldm_2025,t2smark_2025,tagwm_2025} preserve the overall design while proposing algorithmic and structural refinements to tackle several limitations. 
In this work, we concentrate on Gaussian Shading style watermarking and primarily compare against the representative baseline Gaussian Shading~\cite{gaussianshading_2024}.

\begin{figure*}[]
  \centering
    \includegraphics[width=\linewidth]{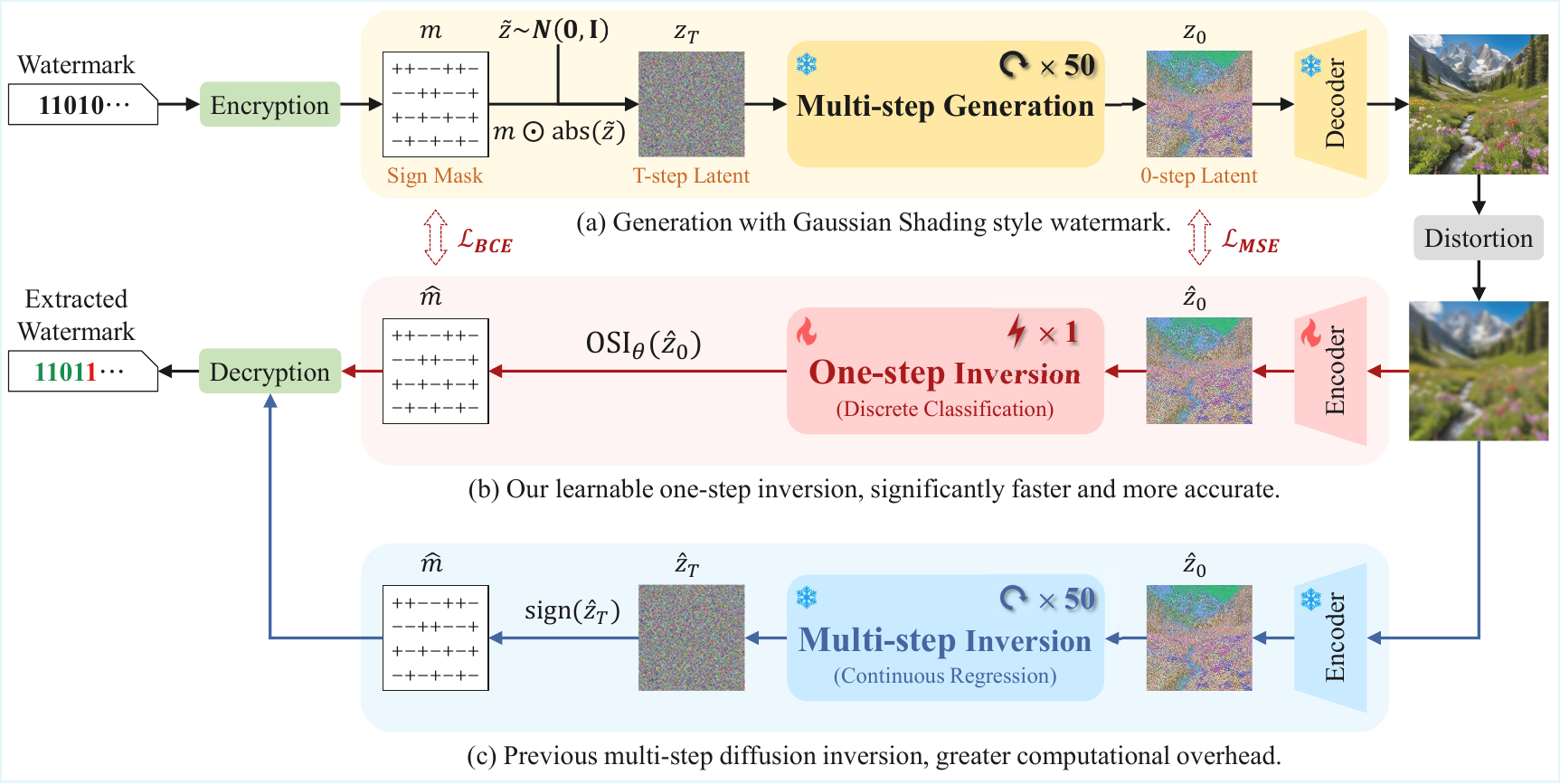}
    \caption{Overview of the proposed OSI method. For generation, (a) we embed Gaussian Shading style watermark, concealing watermark in the signs of initial latent. For extraction, (b) we adopt learnable one-step inversion (OSI) to directly predict the sign mask, while (c) previous methods first perform multi-step diffusion inversion to reconstruct continuous latent $\hat{z}_T$ and then extract the sign of $\hat{z}_T$.}
  \label{fig:overview}
\end{figure*}

\section{Methods}
In \cref{subsec:gs}, we review the baseline, Gaussian Shading~\cite{gaussianshading_2024}.
In \cref{subsec:OSI}, we introduce our One-step Inversion (OSI), a learning-based method that significantly improves the extraction of Gaussian Shading style watermarks.
In \cref{subsec:dataset}, we describe the construction of our training data.

\subsection{Preliminary: Gaussian Shading (GS)}
\label{subsec:gs}
\cref{fig:overview}(a) and \cref{fig:overview}(c) illustrate the Gaussian Shading (GS) watermarking system, which consists of three steps: 1) watermark encryption, 2) diffusion generation, and 3) watermark extraction.

\paragraph{Watermark Encryption} 
GS employs repetition coding and XOR/ChaCha20~\cite{chacha_2008} to embed the watermark information into a sign mask.
Specifically, the binary watermark $wm\in\{0,1\}^{c\times \frac{h}{f_{hw}}\times \frac{w}{f_{hw}}}$ is first replicated along the height and width dimensions to match the latent shape $c\times h\times w$ of the diffusion model, where $f_{hw}$ is the repetition factor for both height and width. 
Then, the repeated watermark is encrypted using XOR or ChaCha20~\cite{chacha_2008} to generate sign mask $m\in \{-1,+1\}^{c\times h\times w}$.

\paragraph{Diffusion Generation with GS Watermarks}
Given the sign mask $m$, the diffusion generation process with GS style watermark is formulated as follows,
\begin{equation}
     I=\mathrm{Dec}(f_{0\leftarrow1}\circ f_{1\leftarrow2}\circ \cdots \circ f_{T-1\leftarrow T}(\underline{\mathrm{abs}(\tilde{z})\odot m})),
     \label{eq:decouple2}
\end{equation}
where $f_{s\leftarrow t}$ denotes the latent diffusion model~\cite{ldm_2022}, $\mathrm{Dec}$ denotes the VAE decoder, and $\tilde{z}\!\sim\!\mathcal{N}(\bm{0},\mathbf{I})$ denotes a Gaussian noise.
As illustrated by $\mathrm{abs}(\tilde{z})\odot m$ in \cref{eq:decouple2}, GS conceals the watermark information as the signs of the initial noise.

\paragraph{Multi-step Inversion for Watermark Extraction}
As can be seen from \cref{eq:decouple2}, the core problem of watermark extraction is to recover the signs $m$ of the initial noise.
To this end, GS first employs multi-step diffusion inversion ~\cite{ddim_2021} to recover the precise initial noise and then extracts the signs of the recovered noise, formulated as follows,
\begin{equation}
    \!\hat{m}=\mathrm{\underline{sign}}(f_{T\leftarrow T-1}\circ f_{T-1\leftarrow T-2}\circ \cdots \circ f_{1\leftarrow 0}(\mathrm{Enc}(I))),
    \label{eq:gsextraction}
\end{equation}
where the inversion step $T$ is typically set as 50.
As a consequence, the watermark extraction requires 50 forward passes of the network $f_{t\leftarrow s}$, which is computationally expensive and time-consuming.

\subsection{One-step Inversion for Watermark Extraction}
\label{subsec:OSI}
Refocusing on the watermark extraction task itself, what we ultimately care about is the signs of the initial noise, rather than its precise value.
What if we directly predict the signs, skipping the cumbersome inversion for exact initial noise?
To this end, we propose One-step Inversion (OSI) for efficient extraction of GS style watermarks.
As shown in \cref{fig:overview}(b), our OSI learns a sign classification model that predicts the signs directly in a single forward pass, formulated as follows,
\begin{align}
    \hat{m}=\mathrm{sign} (p-0.5),\quad p=\mathrm{OSI}_\theta(\mathrm{Enc}_\psi(I)),
    \label{eq:osiextraction}
\end{align}
where $\mathrm{Enc}_\psi$ is a trainable encoder that maps the watermarked image back to the latent space, $\mathrm{OSI}_\theta$ is a trainable sign classification model, and $p$ denotes the predicted probability of the positive sign.

To optimize the model parameters, we employ binary cross-entropy loss for sign classification, as follows,
\begin{equation}
    {\mathcal{L}_\mathrm{BCE}=-y\log(p)-(1-y)\log(1-p)},
    \label{eq:bce}
\end{equation}
where $y=(m+1)/2$ denotes the ground-truth label.
In addition, we employ MSE loss between the encoded latent $\hat{z}_0=\mathrm{Enc}_\psi(I)$ and the original latent $z_0$ (see \cref{fig:overview}(a)), enforcing the latent-consistency to further improve the performance, as follows,
\begin{equation}
    \mathcal{L}_\mathrm{MSE}=\|\hat{z}_0-z_0\|^2_2.
    \label{eq:mse}
\end{equation}
Overall, we train our OSI model with the objective:
\begin{equation}
    \mathop{\min}\limits_{\theta,\psi}\mathcal{L}_\mathrm{BCE}+\mathcal{L}_\mathrm{MSE}.
    \label{eq:objective}
\end{equation}

\begin{figure*}[]
    \centering
    \includegraphics[width=1\linewidth]{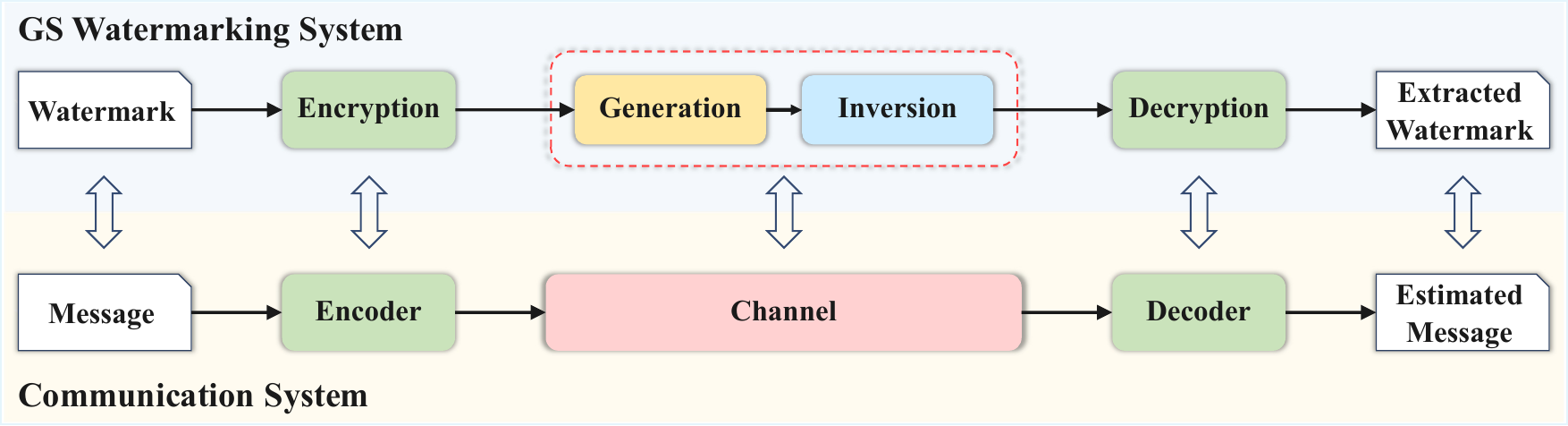}
    \caption{A communication system perspective on Gaussian Shading watermarking systems.\vspace{2ex}}
    \label{fig:communication}
\end{figure*}
Besides, we initialize $\mathrm{OSI}_\theta$ with the pretrained latent diffusion backbone and initialize $\mathrm{Enc}_\psi$ with the pretrained VAE encoder to leverage their inherent capability of exact noise inversion~\cite{ddim_2021,gaussianshading_2024}, which we found to significantly benefit the training convergence.
We also introduce data augmentation, e.g., resizing, blurring, and cropping, to the input images during training, which enhances the model's robustness to various distortions.

\subsection{Training Data Construction}
\label{subsec:dataset}
To train our OSI model, we construct a supervised dataset where each sample is a triplet $(I,z_0,z_T)$.
Specifically, we 1) sample an initial noise $z_T\!\sim\!\mathcal{N}(\bm{0},\mathbf{I})$ whose signs serve as the ground-truth $m$,
2) run the diffusion generation pipeline to obtain the 0-step latent $z_0$ that is used to calculate $\mathcal{L}_\mathrm{MSE}$ in \cref{eq:mse},
and 3) decode $z_0$ to obtain the synthesized image $I$ that serves as the model input.

\section{Communication System Perspective}
\label{sec:communication}
In this section, we reinterpret the GS style watermarking from the perspective of \textbf{communication system}.
As illustrated in \cref{fig:communication}, the watermark can be viewed as the message, the encryption and decryption modules can be viewed as the encoder and the decoder, and the generation-inversion process corresponds to a channel.
From this perspective, we can decouple the watermarking system into two essential components: 1) the \textbf{channel} that transmits the watermark information and 2) the \textbf{coding} scheme that determines the way to encode and decode the watermarks.

According to Shannon's theorem~\cite{shannon_1948}, every channel has an inherent channel capacity $C$, which serves as the theoretical upper bound for the reliable information rate $R$, i.e., $R < C$.
Moreover, there exists a coding scheme such that $R$ can approach $C$ arbitrarily closely, and the error probability can be made arbitrarily small.
Therefore, this principle implies two avenues for optimizing a watermarking system: 1) improving the coding methodology to enhance the information reliability and maximize the information rate $R$ toward the channel capacity $C$, and 2) improving the channel itself to increase the channel capacity.
From this perspective, we can classify existing GS style watermarking methods, including ours, into the following two categories.

\paragraph{Prior Works: Coding Improvement}
Most existing GS style approaches can be regarded as improving the coding scheme over a fixed channel, as they focus on the encryption module while maintaining the generation and inversion procedures unchanged.
For example, GS~\cite{gaussianshading_2024} uses repetition codes (REP), PRCW~\cite{prcwatermark_2025} uses pseudo-random codes (PRC)~\cite{prcode_2024}, and GS++~\cite{gaussianshading++_2025} combines REP and PRC.
Although these coding-centric approaches have made substantial advancements in enhancing both watermark reliability and payload capacity, their performance remains fundamentally constrained by the fixed channel capacity.

\paragraph{Our Work: Channel Improvement}
From the perspective of communication system, our approach can be viewed as an improvement of the channel.
By explicitly training the inversion model to achieve higher bit accuracy, we effectively increase the channel capacity $C$, thereby improving the overall reliability of watermark extraction, as demonstrated in \cref{subsec:sd21} and \cref{subsec:broaderapplication}. 
Notably, several prior coding-centric studies~\cite{gaussianshading_2024,gaussianshading++_2025} have highlighted the need for more accurate diffusion inversion techniques ~\cite{edict_2023,onexact_2024,gradientfreedecoderinversion_2024} to enhance watermark extraction. From the perspective of communication system, the pursuit of these advanced diffusion inversion techniques can be interpreted as striving for more robust and efficient channels.

\paragraph{}
Based on the preceding analysis, we have demonstrated a clear connection between the GS watermarking system and the communication system.
The latter has evolved over many years and is characterized by robust theories and advanced techniques.
In this light, certain research challenges within the GS watermarking system may be reformulated in terms of communication system paradigms, thereby enabling the adoption of mature theories and techniques from communication system.
We hope that this communication system perspective will offer broader and novel insights into the GS watermarking system and promote further advancements in future research.

\begin{table*}[]
    \centering
    \small
    \caption{Evaluation results on SD2.1. ``Adv.'' denotes the average performance in all adversarial scenarios. Across all $f_{hw}$ settings, OSI outperforms GS~\cite{gaussianshading_2024} in TPR, bit accuracy, and payload rate, while substantially reducing FLOPs by over $20\times$ and achieving $25\times$ speedup.}
    {
    \input{tabular/sd21_overview}
    }
    
    \label{tab:sd21_overview}
\end{table*}

\section{Experiment}
In \cref{subsec:implementation}, we introduce implementation details. 
In \cref{subsec:sd21}, we present experimental results of OSI on SD2.1, which shows superiority in both performance and efficiency.
In \cref{subsec:ablation}, we provide detailed ablation studies. 
In \cref{subsec:broaderapplication}, we demonstrate the broad generality of OSI across diffusion models, schedulers, and cryptographic schemes.

\subsection{Implementation detail}
\label{subsec:implementation}
\paragraph{Diffusion Models} 
We focus on text-to-image latent diffusion models and select Stable Diffusion (SD)~\cite{ldm_2022} to conduct experiments. 
The majority of experiments run with SD2.1~\cite{ldm_2022} to align with GS~\cite{gaussianshading_2024}, producing $512\times512$ images from $4\times64\times64$ latents.
To assess generality, we also conduct experiments on SDXL~\cite{sdxl_2024} for higher resolution ($1024\times 1024$ images and $4\times 128\times 128$ latents), and SD3.5~\cite{sd3_2024} for flow matching~\cite{flowmatching_2023,flowstraight_2023} mechanism and DiT~\cite{dit_2023} backbone ($1024\times 1024$ images and $16\times 128\times 128$ latents). 
For all SD variants we adopt classifier-free guidance (CFG)~\cite{cfg_2022}, using a guidance scale of 7.5 during generation and disabling it during inversion (scale set to 1).

\paragraph{Schedulers} 
Following GS~\cite{gaussianshading_2024}, for SD2.1 and SDXL, we sample with 50-step DPM-Solver~\cite{dpmsolver_2022} and inverse with 50-step DDIM scheduler~\cite{ddim_2021}.
For SD3.5, we use the default Euler–discrete scheduler~\cite{sd3_2024} with 28 steps of sampling and inversion.
Note that the inversion schedulers apply only to the baselines. Our OSI inverses one step without scheduler.

\paragraph{Dataset}
We use the Stable Diffusion Prompts (SDP) corpus~\cite{sdp_2024}. Following GS~\cite{gaussianshading_2024}, the first 1k prompts are reserved for evaluation, while the remaining $\sim$72k prompts are used to synthesize the noise–image pairs described in \cref{subsec:dataset}.
To assess cross-dataset generality, we additionally evaluate on prompts from MS-COCO~\cite{mscoco_2014}.

\paragraph{Finetuning Setting}
We finetune the OSI model for 11 epochs with a batch size of 16 using Adam~\cite{adam_2015} and a learning rate of $1\times10^{-4}$.
Random data augmentation (cropping, blurring, etc.) is applied with probability 50\%.

\paragraph{Evaluation Metrics}
Following GS ~\cite{gaussianshading_2024}, we report 1)  the true positive rate at a fixed false positive rate of $10^{-6}$ (TPR@FPR=1e-6; detailed in the Appendix) and 2) bit accuracy of the recovered watermark. 
To assess robustness, we also adopt the nine representative adversarial distortions used in GS~\cite{gaussianshading_2024} (see Appendix).
To quantify efficiency, we measure FLOPs and the payload rate (see \cref{para:payload}).
Note that we only modify the extraction pipeline relative to prior methods~\cite{gaussianshading_2024,prcwatermark_2025,t2smark_2025}.
Therefore, image quality metrics (e.g., FID~\cite{fid_2017} and CLIP-Score~\cite{clip_2021}) are expected to be essentially identical and thus are not considered in our study.

\subsection{Performance of One-step inversion (OSI)}
\paragraph{Performance Gain}
In \cref{tab:sd21_overview}, we compare our OSI with the GS~\cite{gaussianshading_2024} baseline under four different $f_{hw}$ settings on SD2.1. 
Generally, larger $f_{hw}$ values yield higher bit accuracy but correspond to shorter embedded watermarks.
Across all $f_{hw}$ configurations, OSI consistently outperforms GS, achieving higher bit accuracy and TPR in both clean and adversarial scenarios.
The cross-dataset evaluations on MS-COCO~\cite{mscoco_2014} exhibit the same performance trends, as shown in \cref{tab:coco}, indicating that the advantages of OSI are not tied to a specific training distribution and transfer robustly to a different domain.
\begin{table}[]
\centering
\small
    \caption{Comparison with GS~\cite{gaussianshading_2024} on MS-COCO~\cite{mscoco_2014} dataset.}
    
    {
        \input{tabular/sd21_coco}
    }
    
    \label{tab:coco}
\end{table}

\paragraph{Computational Efficiency}
As shown in \cref{tab:sd21_overview}, OSI reduces the computational cost by more than $20\times$ in terms of FLOPs. To further quantify the practical benefit, we also measure the per-extraction wall-clock time on a single NVIDIA A100 40G GPU. The results in \cref{tab:sd21_overview} corroborate the efficiency gain, showing more than $25\times$ speedup in runtime, which confirms that OSI delivers substantial gains in both computation and actual inference latency.

\paragraph{Amortized Training Cost at Scale}
Although the improvements of OSI are obtained at the expense of finetuning, this cost is negligible when amortized over downstream deployment at scale. 
The finetuning procedure described in \cref{subsec:implementation} requires only about 15 hours on 8 NVIDIA A100 40G GPUs, corresponding to roughly $1.92\mathrm{T}\times11\times72\mathrm{k}\approx 1.52\mathrm{E}$ FLOPs in total. 
Given that OSI saves about $39\mathrm{T}$ FLOPs per extraction run, only $1.52\mathrm{E}\div 39\mathrm{T}\approx39\mathrm{k}$ extraction runs are needed to fully amortize the finetuning compute, which is well within the scale of platform-level deployments. 
Moreover, as shown in \cref{subsec:broaderapplication}, the finetuned OSI model generalizes across diverse generation schedulers and cryptographic schemes, allowing a single finetuning investment to support a broad range of scenarios.
Overall, this one-time finetuning expense yields a highly favorable return on compute and is amortized rapidly at scale.

\begin{figure}[]
  \centering
    \includegraphics[width=1\linewidth]{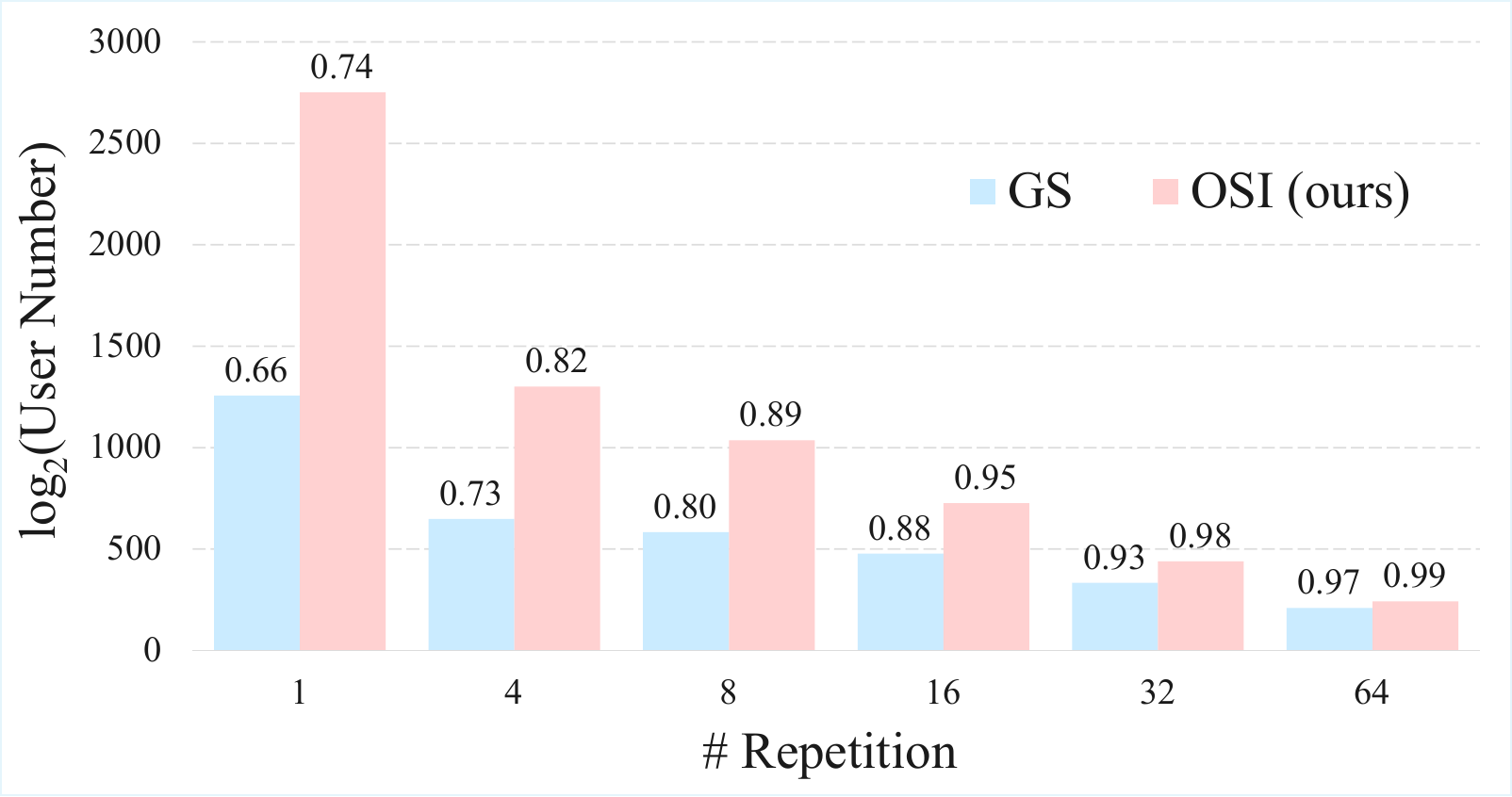}
  \caption{Comparison of user number across repetition count settings. Numbers above the bars display the corresponding bit accuracies. The higher accuracy of OSI yields an exponentially larger addressable user number than GS~\cite{gaussianshading_2024}.}
  \label{fig:payload_acc}
\end{figure}
\label{subsec:sd21}
\paragraph{Payload Rate}\label{para:payload}
From a communication-theoretic perspective, we analyze the payload rate $R$ of GS and OSI, which quantifies the amount of effective watermark information that can be embedded per position in $z_T$.
We model the watermarking-extraction procedure as a binary symmetric channel (BSC) with crossover probability $p=1-a$, where $a$ is the bit accuracy reported in \cref{tab:sd21_overview}. According to Shannon's theorem~\cite{shannon_1948}, the channel capacity of this BSC is 
\begin{equation}
    \begin{aligned}
        C_{\mathrm{BSC}} &= 1 - H(p) \\
        &= 1+p\log_2 p+(1-p)\log_2(1-p),
    \end{aligned}
    \label{eq:bsc-capacity}
\end{equation}
which characterizes the maximum proportion of effective payload bits in communication. Considering the $f_{hw}^2$ replication used in GS, the payload rate $R$ is calculated as 
\begin{equation}
    R=\frac{C_\mathrm{BSC}}{f_{hw}^2}.
    \label{eq:coderate}
\end{equation}
Thus, when using an error-correcting code (ECC) that operates close to the capacity $C_\mathrm{BSC}$, the maximum reliably embedded payload in the latent $z_T$ is $R\times c\times h\times w$ bits. Notably, when $f_{hw}=1$, OSI avoids replication (no coding involved) and achieves a $2\times$ higher payload rate than GS, indicating OSI can consistently outperform GS under the same coding scheme.

For an intuitive comparison between OSI and GS, we use all effective payload bits to encode a user identifier, which allows for $2^{R\times c\times h\times w}$ distinct users in total. We visualize the resulting gap in the user number between OSI and GS in \cref{fig:payload_acc}. By improving bit accuracies, OSI achieves higher payload rates $R_\mathrm{OSI} > R_\mathrm{GS}$, thereby supporting exponentially larger user numbers than GS.

\begin{table}[]
    \centering
    \small
    \caption{Performance under advanced attacks.}
    \resizebox{1.0\linewidth}{!}{%
        \input{tabular/advanced_attack}
    }
    \label{tab:advanced attack}
\end{table}

\paragraph{Advanced Attack}
We introduce two types of advanced attacks to evaluate the performance of OSI under more rigorous conditions. 
The first type is compression attack, where pretrained VAEs ~\cite{bmshj2018,mbt2018,cheng2020} are used to encode and decode watermarked images to emulate realistic lossy compression in practical deployment.
The second type is embedding attack~\cite{featurespaceperturbation_2019}, which generates perturbations that are nearly imperceptible in pixel space but significantly different in feature embedding space (ResNet/CLIP)~\cite{resnet_2016, clip_2021}. In this way, embedding attacks can induce extraction failures, particularly for watermarking methods that operate at the latent or feature level.
As summarized in \cref{tab:advanced attack}, OSI consistently outperforms GS under both attack types.

\begin{table*}[]
    \centering
    \small
    \caption{Module ablation on SD2.1. We compare different settings without data augmentation. Across all $f_{hw}$ configurations, jointly finetuning the UNet and encoder achieves the best performance in both clean and adversarial scenarios.}
    \input{tabular/module_ablation}
    
    \label{tab:module ablation}
\end{table*}

\subsection{Ablation Study}
\label{subsec:ablation}
\paragraph{OSI Model Design}
\label{subsubsec:module ablation}
The default OSI model comprises the encoder and the diffusion backbone model (UNet of SD2.1) of the generation pipeline, leveraging their watermark extraction capability demonstrated in multi-step diffusion inversion~\cite{gaussianshading_2024,prcwatermark_2025}. 
A natural question is whether these components can be directly utilized for one-step inversion. 
We first evaluate a fully frozen configuration, fixing the encoder and the U-Net and performing one-step extraction. 
The ``Frozen'' result in \cref{tab:module ablation} shows certain one-step inversion capability, but performance lags markedly behind the GS baseline, indicating that finetuning is necessary.
We then consider two fine-tuning settings: 1) freezing the encoder and finetuning only the U-Net; and 2) finetuning both the encoder and the U-Net. 
Both settings outperform the GS baseline, with the best results obtained when finetuning both modules. 
Accordingly, we adopt joint finetuning as the default configuration.

\begin{figure}[]
  \centering
    \includegraphics[width=1\linewidth]{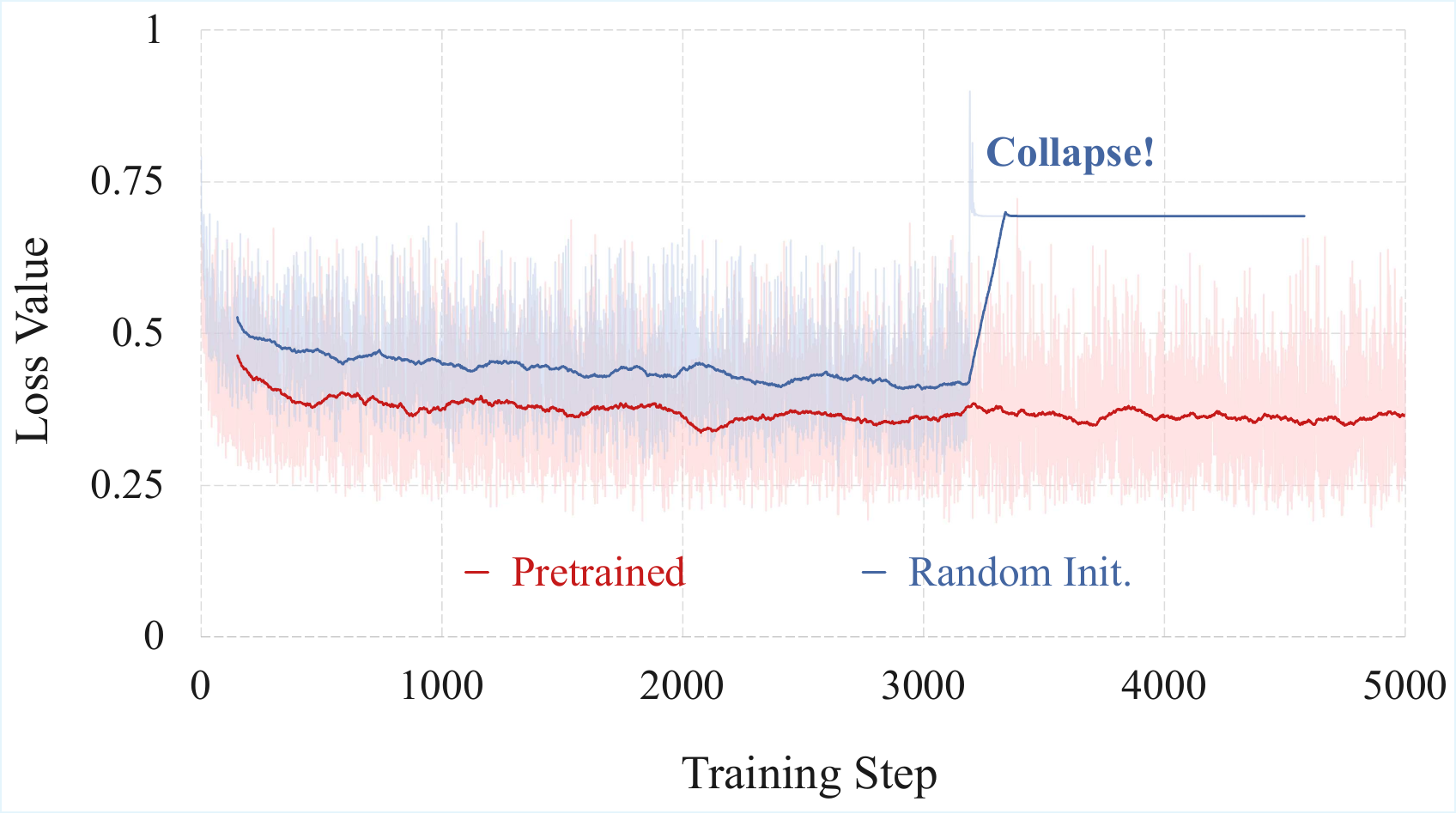}
  \caption{Training loss comparison between different initializations of UNet parameters. Random initialization encounters training collapse and fails to converge.}
  \label{fig:unetinit}
\end{figure}

Conceptually, we formulate the watermark extraction as a classification problem, any sufficiently capable classifier could be used.
Why not train a lightweight model from scratch? 
In practice, finetuning the pretrained diffusion modules is more reliable: it leverages features already aligned with the latent space, stabilizes optimization, and avoids the overhead of crafting new architectures and training schedules. 
As a reference point, we randomly initialize a U-Net and train it under identical settings. 
As shown in \cref{fig:unetinit}, the training of random initialization collapses and fails to converge. 
Collectively, these analyses and observations confirm that the initialization with the parameters of the encoder and diffusion backbone model is a sensible and effective choice for our OSI model.

\paragraph{Data Augmentation}
With probability 0.5, we apply a random type of image distortion to the inputs during finetuning to improve the robustness of OSI. 
The data augmentation is not tailored to the evaluation protocol. Instead, it helps OSI tolerate a broader range of image distortions.
As shown in \cref{tab:data aug}, with $f_{hw}=1$, finetuning without data augmentation already exceeds GS baseline, while adding augmentation further boost performances in adversarial scenarios, with only a minor drop in the clean scenario. 
\begin{table}[]
    \centering
    \small
    \caption{Performance of OSI with and without data augmentation.}
    \input{tabular/data_aug}
    
    \label{tab:data aug}
\end{table}

\paragraph{Classifier-Free Guidance (CFG)}
We set the CFG scale to 7.5 when synthesizing our dataset for finetuning the OSI model.
Considering the variety of CFG scales applied during generation, we assess the performance of OSI under different guidance scales with $f_{hw}=1$ in \cref{tab:cfg}. 
OSI consistently outperforms GS across all tested CFG settings, despite being trained solely on noise–image pairs which are generated with CFG scale of 7.5.

\begin{table}[]
\small
    \caption{Performances across different CFG scales.}
    \resizebox{1.0\linewidth}{!}{%
        \input{tabular/cfg}
    }
    \label{tab:cfg}
\end{table}

\subsection{Broader Application}
\label{subsec:broaderapplication}
\paragraph{SDXL}
We extend OSI to SDXL~\cite{sdxl_2024} to test performance in high-resolution generation.
Analogous to SD 2.1, we synthesize a noise–image dataset at corresponding shapes and finetune OSI using the default configuration in \cref{subsubsec:module ablation}.
As shown in \cref{tab:sdxl}, OSI mirrors the trends observed on SD2.1: it delivers higher accuracy and payload rate while requiring substantially fewer FLOPs.
\begin{table}[]
\caption{Comparison with GS~\cite{gaussianshading_2024} on SDXL.}
    \resizebox{1.0\linewidth}{!}{%
        \input{tabular/sdxl}
    }
    \label{tab:sdxl}
\end{table}

\paragraph{SD3.5}
We also extend OSI to SD3.5~\cite{sd3_2024}, a model based on the flow matching~\cite{flowmatching_2023,flowstraight_2023} mechanism and the DiT architecture~\cite{dit_2023}.
We synthesize the dataset and fine-tune OSI as before.
The results reported in \cref{tab:sd35} again demonstrate that OSI outperforms GS in performance and efficiency.
\begin{table}[]
\caption{Comparison with GS~\cite{gaussianshading_2024} on SD3.5.}
    \resizebox{1.0\linewidth}{!}{%
        \input{tabular/sd35}
    }
    \label{tab:sd35}
\end{table}

\begin{table}[]
    \centering
    \caption{Comparison of different scheduler pairs on clean images.}
    \resizebox{1.0\linewidth}{!}{%
    \input{tabular/scheduler}
    }
    \label{tab:scheduler}
\end{table}

\paragraph{Scheduler}
We assess OSI using three (generation | inversion) scheduler configurations: (DPM-Solver~\cite{dpmsolver_2022} | DPM-Solver), (DPM-Solver | DDIM~\cite{ddim_2021}), and (DDIM | DDIM).
Notably, OSI only depends on the generation scheduler, whereas GS relies on both generation scheduler and inversion scheduler.
The performance of the three combined scheduler settings in \cref{tab:scheduler} ($f_{hw}=1$) shows that OSI consistently outperforms GS across all settings with higher bit accuracy and significantly lower FLOPs.

\begin{table}[]
\small
\caption{Comparison with PRC~\cite{prcwatermark_2025} on SD2.1.}
\resizebox{1.0\linewidth}{!}{%
\input{tabular/prc}
}
    
    \label{tab:prc}
\end{table}

\paragraph{PRCW}
Since OSI does not alter the cryptography operations, it can be applied to other GS style methods that adopt different codes.
Here we apply OSI to PRCW~\cite{prcwatermark_2025}, which leverages the PRC~\cite{prcode_2024} code. 
We integrate OSI with PRCW~\cite{prcwatermark_2025} by replacing its multi-step diffusion inversion with the OSI model. 
As shown in \cref{tab:prc}, across all settings of watermark bits, OSI achieves higher bit accuracy.

\paragraph{T2SMark}
We further evaluate OSI on T2SMark~\cite{t2smark_2025}, a recent method which follows the GS framework but introduces tail-truncated sampling in the cryptographic stage.
After substituting the original multi-step inversion with the finetuned OSI model, we achieve higher bit accuracy across all settings in \cref{tab:t2smark}, confirming the superiority of OSI.
\begin{table}[]
    \caption{Comparison with T2SMark~\cite{t2smark_2025} on SD2.1.}
    \resizebox{1.0\linewidth}{!}{%
    \input{tabular/t2s}
    } 
    \label{tab:t2smark}
\end{table}

\section{\vspace{-0.3ex}Conclusion}
We revisit Gaussian Shading style watermarking from communication system perspective and reformulate watermark extraction as a discrete classification problem rather than a reduced form of continuous latent regression. 
Building on these views, we propose One-step Inversion (OSI), a learnable one-step watermark extractor that serves as a channel improvement to the GS framework. 
Extensive experiments across various schedulers, diffusion models, and cryptographic schemes demonstrate that OSI consistently offers stronger extraction performance, higher payload capacity, and substantially lower computational cost compared with existing approaches. 
We hope our work and perspective provide practical and principled references for developing more efficient and robust diffusion watermarking systems.
\newpage

%% file: tabular/sd21_overview.tex
\begin{tabularx}{\textwidth}{>{\centering\arraybackslash}m{1.15cm}>{\centering\arraybackslash}m{0.55cm}>{\centering\arraybackslash}X
>{\centering\arraybackslash}X>{\centering\arraybackslash}X>{\centering\arraybackslash}X>{\centering\arraybackslash}X>{\centering\arraybackslash}X>{\centering\arraybackslash}X}
\toprule
\multirow{2}{*}[-0.7ex]{Method}                                               & \multirow{2}{*}[-0.7ex]{$f_{hw}$} & \multicolumn{2}{c}{TPR@FPR=1e-6}  & \multicolumn{2}{c}{Bit Accuracy} & \multirow{2}{*}[-0.7ex]{\begin{tabular}[c]{@{}c@{}}Payload\\Rate\end{tabular}} & \multirow{2}{*}[-0.7ex]{FLOPs (T)} & \multirow{2}{*}[-0.7ex]{A100 Time (s)}    \\ \cmidrule(lr){3-4} \cmidrule(lr){5-6} 
                                                                      &                           & Clean      & Adv.         & Clean           & Adv.     &                               &                                \\ \midrule
\multirow{4}{*}{GS}                                                   & 1                         & 1.0000     & 0.9982       & 0.8560          & 0.6616          & 7.67\%                        & \multirow{4}{*}{41.3} & \multirow{4}{*}{1.52}          \\
                                                                      & 2                         & 1.0000     & 0.9967       & 0.9408          & 0.7298          & 3.96\%                        & &                               \\
                                                                      & 4                         & 1.0000     & 0.9964       & 0.9984          & 0.8784          & 2.91\%                        & &                               \\
                                                                      & 8                         & 1.0000     & 0.9970       & 1.0000          & 0.9728          & 1.28\%                        & &                               \\ \midrule
\multirow{4}{*}{\begin{tabular}[c]{@{}c@{}}OSI\\(ours)\end{tabular}} & 1                         & 1.0000     & \textbf{0.9991}       & \textbf{0.8839} & \textbf{0.7364} & \textbf{16.8\%}               & \multirow{4}{*}{\textbf{1.92}} & \multirow{4}{*}{\textbf{0.06}} \\
                                                                      & 2                         & 1.0000     & \textbf{0.9984}       & \textbf{0.9607} & \textbf{0.8189} & \textbf{7.94\%}               & &                               \\
                                                                      & 4                         & 1.0000     & \textbf{0.9986}       & \textbf{0.9996} & \textbf{0.9491} & \textbf{4.44\%}               & &                               \\
                                                                      & 8                         & 1.0000     & \textbf{0.9984}       & 1.0000          & \textbf{0.9939} & \textbf{1.48\%}               & &                               \\ 
\bottomrule
\end{tabularx}

%% file: tabular/sd21_coco.tex
\begin{tabularx}{\linewidth}{>{\centering\arraybackslash}m{1.15cm}>{\centering\arraybackslash}m{0.55cm}>{\centering\arraybackslash}X>{\centering\arraybackslash}X>{\centering\arraybackslash}m{1.9cm}}
    \toprule
    Method & $f_{hw}$ & Clean & Adv. & Payload Rate \\ \midrule
    \multirow{4}{*}{GS} 
     & 1 & 0.8652 & 0.6616 & 7.67\% \\
     & 2 & 0.9473 & 0.7290 & 3.93\% \\
     & 4 & 0.9992 & 0.8760 & 2.87\% \\
     & 8 & 1.0000 & 0.9697 & 1.26\% \\ \midrule
    \multirow{4}{*}{\begin{tabular}[c]{@{}c@{}}OSI\\(ours)\end{tabular}}
     & 1 & \textbf{0.8815} & \textbf{0.7326} & \textbf{16.2\%} \\
     & 2 & \textbf{0.9590} & \textbf{0.8134} & \textbf{7.64\%} \\
     & 4 & \textbf{0.9997} & \textbf{0.9445} & \textbf{4.32\%} \\
     & 8 & 1.0000 & \textbf{0.9931} & \textbf{1.47\%} \\ 
     \bottomrule
\end{tabularx}

%% file: tabular/advanced_attack.tex
\begin{tabular}{cccccc}
    \toprule
    \multirow{2}{*}{} & \multirow{2}{*}{} & \multicolumn{4}{c}{$f_{hw}$} \\ \cmidrule(lr){3-6}
    \multicolumn{2}{c}{} & 1 & 2 & 4 & 8 \\ \midrule
    \multirow{2}{*}{Compression} 
     & GS & 0.6253 & 0.6822 & 0.8264 & 0.9460 \\
     & OSI & \textbf{0.6339} & \textbf{0.6941} & \textbf{0.8434} & \textbf{0.9584}  \\ \midrule
    \multirow{2}{*}{Embedding} 
     & GS & 0.6394 & 0.7003 & 0.8585 & 0.9647 \\
     & OSI & \textbf{0.6483} & \textbf{0.7188} & \textbf{0.8804} & \textbf{0.9758} \\ 
     \bottomrule
\end{tabular}

%% file: tabular/module_ablation.tex
\begin{tabularx}{\textwidth}{l >{\centering\arraybackslash}X>{\centering\arraybackslash}X>{\centering\arraybackslash}X>{\centering\arraybackslash}X>{\centering\arraybackslash}X>{\centering\arraybackslash}X>{\centering\arraybackslash}X>{\centering\arraybackslash}X}
    \toprule
    \multirow{2}{*}[-0.7ex]{Method} & \multicolumn{4}{c}{Clean}                                                        & \multicolumn{4}{c}{Adversarial}                                       \\ \cmidrule(lr){2-5} \cmidrule(lr){6-9}  
                            & 1               & 2               & 4               & \multicolumn{1}{c}{8}      & 1               & 2               & 4               & 8               \\ \midrule
    Baseline (GS)                      & 0.8560          & 0.9408          & 0.9984          & \multicolumn{1}{c}{1.0000} & 0.6616          & 0.7298          & 0.8784          & 0.9728          \\
    Frozen                  & 0.7241          & 0.8119          & 0.9612          & \multicolumn{1}{c}{0.9985} & 0.6148          & 0.6682          & 0.8041          & 0.9316          \\
    Tune UNet               & 0.8810          & 0.9507          & 0.9991          & \multicolumn{1}{c}{1.0000} & 0.6664          & 0.7342          & 0.8831          & 0.9749          \\
    Tune UNet \& Encoder (OSI)     & \textbf{0.8865} & \textbf{0.9625} & \textbf{0.9997} & \multicolumn{1}{c}{1.0000} & \textbf{0.6729} & \textbf{0.7445} & \textbf{0.8953} & \textbf{0.9809} \\ 
    \bottomrule
    
\end{tabularx}

%% file: tabular/data_aug.tex
\begin{tabularx}{\linewidth}{>{\centering\arraybackslash}X>{\centering\arraybackslash}X>{\centering\arraybackslash}X>{\centering\arraybackslash}X}
    \toprule
     & GS & OSI w/o & OSI w/ \\ \midrule
    Clean & 0.8560 & \textbf{0.8865} & 0.8839 \\
    Adversarial & 0.6616 & 0.6729 & \textbf{0.7364} \\ 
    \bottomrule
\end{tabularx}

%% file: tabular/cfg.tex
\begin{tabular}{ccccccc}
    \toprule
    \multicolumn{2}{c}{\multirow{2}{*}[-0.7ex]} & \multicolumn{5}{c}{CFG} \\ \cmidrule(lr){3-7}
    \multicolumn{2}{c}{} & 1 & 3 & 5 & 7.5 & 10 \\ \midrule
    \multirow{2}{*}{Clean} & GS & 0.9263 & 0.9044 & 0.8826 & 0.8560 & 0.8307 \\
     & OSI & \textbf{0.9362} & \textbf{0.9217} & \textbf{0.9057} & \textbf{0.8839} & \textbf{0.8604} \\ \midrule
    \multirow{2}{*}{Adversarial} & GS & 0.7281 & 0.7007 & 0.6801 & 0.6616 & 0.6474 \\
     & OSI & \textbf{0.7963} & \textbf{0.7738} & \textbf{0.7553} & \textbf{0.7364} & \textbf{0.7199} \\ 
     \bottomrule
\end{tabular}

%% file: tabular/sdxl.tex
\begin{tabular}{cccccc}
    \toprule
    Method & $f_{hw}$ & Clean & Adv. & Payload Rate & FLOPs(T) \\ \midrule
    \multirow{4}{*}{GS} 
     & 1 & 0.7845 & 0.6123 & 3.67\% &  \multirow{4}{*}{303.3} \\
     & 2 & 0.8789 & 0.6649 & 2.00\% & \\
     & 4 & 0.9894 & 0.8008 & 1.75\% & \\
     & 8 & 0.9999 & 0.9305 & 0.99\% & \\ \midrule
    \multirow{4}{*}{\begin{tabular}[c]{@{}c@{}}OSI\\(ours)\end{tabular}} 
     & 1 & \textbf{0.8276} & \textbf{0.6676} & \textbf{8.26}\% &  \multirow{4}{*}{\textbf{10.3}} \\
     & 2 & \textbf{0.9185} & \textbf{0.7395} & \textbf{4.31}\% & \\
     & 4 & \textbf{0.9963} & \textbf{0.8914} & \textbf{3.15}\% & \\
     & 8 & \textbf{1.0000} & \textbf{0.9790} & \textbf{1.33}\% & \\ 
     \bottomrule
\end{tabular}

%% file: tabular/sd35.tex
\begin{tabular}{cccccc}
    \toprule
    Method & $f_{hw}$ & Clean & Adv. & Payload Rate & FLOPs(T)  \\ \midrule
    \multirow{4}{*}{GS} 
     & 1 & 0.7604 & 0.5810 & 1.90\% & \multirow{4}{*}{200.9} \\
     & 2 & 0.8520 & 0.6197 & 1.04\% &  \\
     & 4 & 0.9746 & 0.7278 & 0.97\% &  \\
     & 8 & 0.9982 & 0.8572 & 0.64\% &  \\ \midrule
    \multirow{4}{*}{\begin{tabular}[c]{@{}c@{}}OSI\\(ours)\end{tabular}} 
     & 1 & \textbf{0.7978} & \textbf{0.6347} & \textbf{5.30\%} & \multirow{4}{*}{\textbf{11.4}}\\
     & 2 & \textbf{0.8887} & \textbf{0.6935} & \textbf{2.77\%} &  \\
     & 4 & \textbf{0.9858} & \textbf{0.8359} & \textbf{2.22\%} & \\
     & 8 & \textbf{0.9993} & \textbf{0.9551} & \textbf{1.15\%} & \\ 
     \bottomrule
\end{tabular}

%% file: tabular/scheduler.tex
\begin{tabular}{ccccccc}
    \toprule
    \multirow{2}{*}[-0.7ex]{Model} & \multirow{2}{*}[-0.7ex]{\begin{tabular}[c]{@{}c@{}}Generation\\ Scheduler\end{tabular}} & \multirow{2}{*}[-0.7ex]{\begin{tabular}[c]{@{}c@{}}Inversion\\ Scheduler\end{tabular}} & \multicolumn{2}{c}{Bit Acc.} & \multicolumn{2}{c}{FLOPs(T)} \\
    \cmidrule(lr){4-5} \cmidrule(lr){6-7}
     &  &  & GS & OSI & GS & OSI \\ \midrule
    \multirow{3}{*}{SD2.1}
     & DDIM & DDIM   & 0.8537 & \textbf{0.8805} & 41.3  & \textbf{1.92} \\
     & DPM-Solver & DDIM  & 0.8560 & \textbf{0.8839} & 41.3  & \textbf{1.92} \\
     & DPM-Solver & DPM-Solver & 0.8655 & \textbf{0.8839} & 197.8 & \textbf{1.92} \\ \midrule
    \multirow{3}{*}{SDXL} 
     & DDIM & DDIM   & 0.8524 & \textbf{0.8774} & 303.3 & \textbf{10.3} \\
     & DPM-Solver & DDIM  & 0.7845 & \textbf{0.8276} & 303.3 & \textbf{10.3} \\
     & DPM-Solver & DPM-Solver & 0.6611 & \textbf{0.8276} & 1738  & \textbf{10.3} \\ 
     \bottomrule
\end{tabular}

%% file: tabular/prc.tex
\begin{tabular}{ccccccc}
\toprule
\multicolumn{2}{c}{\multirow{2}{*}{}} & \multicolumn{5}{c}{Watermark Bits} \\ \cmidrule{3-7} 
\multicolumn{2}{c}{} & 4096 & 2048 & 1024 & 512 & 256 \\ \midrule
\multirow{2}{*}{Clean} & PRC & 0.8751 & 0.9480 & 1.0000 & 1.0000 & 1.0000 \\
 & OSI & \textbf{0.8752} & \textbf{0.9605} & 1.0000 & 1.0000 & 1.0000 \\ \midrule
\multirow{2}{*}{Adversarial} & PRC & 0.8751 & 0.9376 & 0.9739 & 0.9903 & 0.9972 \\
 & OSI & 0.8751 & \textbf{0.9395} & \textbf{0.9925} & \textbf{0.9993} & \textbf{0.9997} \\ \bottomrule
\end{tabular}

%% file: tabular/t2s.tex
\begin{tabular}{ccccccc}
\toprule
\multicolumn{2}{c}{\multirow{2}{*}{}} & \multicolumn{5}{c}{Watermark Bits} \\ \cmidrule{3-7} 
\multicolumn{2}{c}{} & 4096 & 2048 & 1024 & 512 & 256 \\ \midrule
\multirow{2}{*}{Clean} 
 & T2S  & 0.9909 & 0.9979 & 0.9998 & 1.0000 & 1.0000 \\
 & OSI  & \textbf{0.9977} & \textbf{0.9998} & \textbf{1.0000} & 1.0000 & 1.0000 \\ \midrule
\multirow{2}{*}{Adversarial} 
 & T2S  & 0.7993 & 0.8635 & 0.9387 & 0.9756 & 0.9909 \\
 & OSI  & \textbf{0.8943} & \textbf{0.9407} & \textbf{0.9884} & \textbf{0.9980} & \textbf{0.9991} \\ \bottomrule
\end{tabular}

%% file: sec/X_suppl.tex
\clearpage
\setcounter{page}{1}
\maketitlesupplementary

\section{Application Scenario}
We consider the application scenario which involves a service provider, users and infringers. 
The service provider deploys a diffusion generation model on the platform but does not release the source codes or model weights. 
Users access the diffusion model via an API to generate images, which they may subsequently publish or distribute for their own purposes.
Infringers may then appropriate these generated images and falsely claim copyright ownership over them. 
In this scenario, the service provider embeds a watermark at generation time to address two safety objectives: \textbf{detection} and \textbf{traceability}. 

For detection, the service provider embeds a watermark that serves as a provenance marker, indicating that an image was generated by the provider’s model and deterring misappropriation by infringers. 

For traceability, the provider records user-specific identity metadata into the watermark to protect the rights of legitimate users, support dispute resolution, and enable accountable tracing of abusive behavior (e.g., generating malicious or harmful content). 

\section{Evaluation Metrics}
\subsection{Bit Accuracy}
For each watermarked image, we define the bit accuracy as the proportion of matched bits between the embedded $k$-bit watermark $wm\in\{0,1\}^k$ and the extracted watermark $\hat{wm}$:
%
\begin{equation}
    {Acc}(wm,\hat{wm})=\frac{1}{k}\sum_{i=1}^k\mathbf{1}(wm_i=\hat{wm}_i).
\end{equation}

\subsection{TPR@FPR}
We report TRP@FPR for detection purpose, where we introduce a threshold $\tau\in[0,1]$ on the bit accuracy. 
An image is classified as watermarked if ${Acc}(wm,\hat{wm}) > \tau$. 
Different choices of $\tau$ lead to different true positive rates (TPR) and false positive rates (FPR). 

In diffusion watermarking, we typically cannot directly measure the FPR, because we only observe positive (watermarked) samples in practice. 
Following GS~\cite{gaussianshading_2024}, for negative (non-watermarked) samples, we assume that their extracted watermark bits $\hat{wm}_1,\hat{wm}_2,...,\hat{wm}_k$ are independently and identically distributed, where each bit follows a Bernoulli distribution $Ber(0.5)$. 
Under this assumption, the bit accuracy of negative (non-watermarked) samples follows a binomial distribution $B(k,0.5)$. 
Therefore, we can use the binomial distribution to approximate the FPR as
\begin{equation}
    \begin{aligned}
         \mathrm{FPR}(\tau)  &= \mathbb{P}(Acc(wm,\hat{wm}) > \tau) \\
                    &= \frac{1}{2^k}\sum_{i=\tau+1}^{k}\binom{k}{i} \\
                    &= \beta_{1/2}(\tau+1,k-\tau),
    \end{aligned}
    \label{eq:fpr beta}
\end{equation}
where $\beta_{1/2}(\cdot,\cdot)$ is a regularized incomplete beta function. 

Finally, we compute TPR@FPR as follows: 
1) Choose the desired FPR. 
2) Solve \cref{eq:fpr beta} to obtain the corresponding threshold $\tau$. 
3) Evaluate the proportion of watermarked images whose bit accuracies exceed the threshold $\tau$, which corresponds to the empirical TPR, defined as
\begin{equation}
    \mathrm{TPR}=\frac{1}{N}\sum_i^{N}\mathbf{1}(Acc(wm_i, \hat{wm}_i) > \tau).
    \label{eq:supp_tpr}
\end{equation}
In our experiments, we fix FPR=1e-6. The resulting thresholds $\tau$ for all $f_{hw}$ configurations are listed in \cref{tab:supp_threshold} (noted that $k=\frac{c\cdot h\cdot w}{f_{hw}^2}$).
\begin{table}[h]
    \centering
    \small
    \caption{Threshold $\tau$ for different $f_{hw}$ with FPR=1e-6.}
    \input{tabular/supp_tpr_fpr_threshold}
    \label{tab:supp_threshold}
\end{table}

With the above construction, TPR@FPR=1e-6 is computed by thresholding bit accuracy and is therefore a monotonic function of bit accuracy. Consequently, we adopt bit accuracy as our primary evaluation metric and report it in all related tables, while TPR@FPR=1e-6 is only included in \cref{tab:sd21_overview} for reference.

Moreover, to complement this single operating point (FPR=1e-6) and provide a more comprehensive comparison, we further sweep the FPR over a logarithmic grid and compute the Area Under the Receiver Operating Characteristic Curve (AUC-ROC)~\cite{aucroc_1997}, as detailed in \cref{subsec:image distortion}.

\section{Robustness Evaluation}
In this section, we provide additional details on the robustness evaluation of OSI.
\subsection{Image Distortion}
\label{subsec:image distortion}
\begin{figure*}[p]
    \centering
    \includegraphics[width=1.0\linewidth]{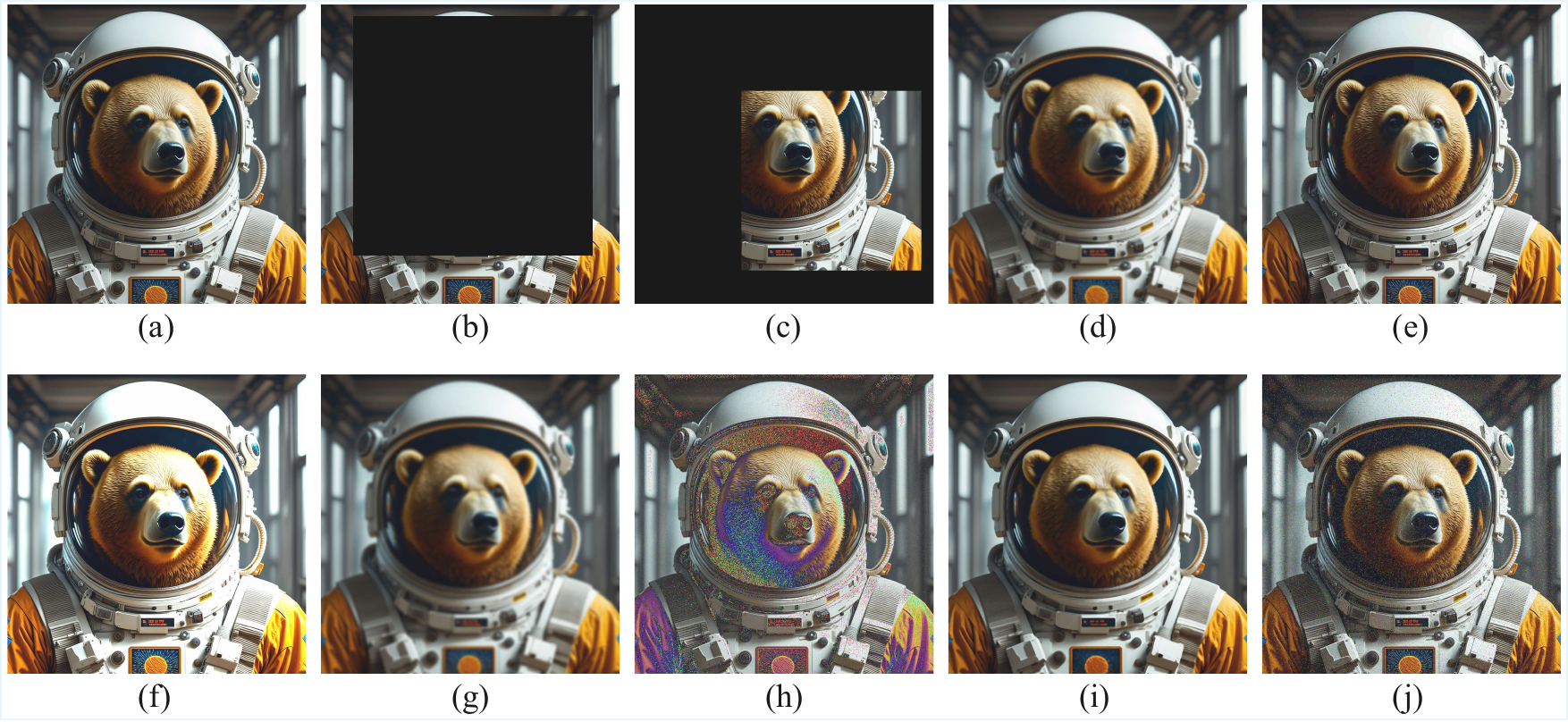}
    \caption{Visualization of different adversarial image distortions. (a) Clean watermarked image. (b). 80\% random drop. (c) 60\% random crop. (d) Resize to 25\% and restore. (e) JPEG with $QF=25$. (f) Brightness with $f=6$. (g) Gaussian blur with $r=4$. (h) Gaussian Noise with $\mu=0,\sigma=0.05$. (i) Median blur with $k=7$. (j) Salt \& Pepper noise with $p=0.05$.}
    \label{fig:103}
\end{figure*}

We adopt nine types of adversarial image distortions to evaluate the robustness of OSI method, as illustrated in \cref{fig:103}. 

\begin{table*}[p]
    \centering
    \caption{Bit accuracy comparisons among all adversarial scenarios with GS~\cite{gaussianshading_2024} on SD2.1.}
    \resizebox{1.0\linewidth}{!}{%
        \input{tabular/supp_bitacc_sd21}
    }
    \label{tab:supp_bitacc_sd21}
\end{table*}

\begin{table*}[p]
    \centering
    \caption{TPR@FPR=1e-6 comparisons among all adversarial scenarios with GS~\cite{gaussianshading_2024} on SD2.1.}
    \resizebox{1.0\linewidth}{!}{%
        \input{tabular/supp_tpr_sd21}
    }
    \label{tab:supp_tpr_sd21}
\end{table*}

\FloatBarrier

\begin{table*}[p]
    \centering
    \caption{Bit accuracy comparisons among all adversarial scenarios with GS~\cite{gaussianshading_2024} on SDXL.}
    \resizebox{1.0\linewidth}{!}{%
        \input{tabular/supp_bitacc_sdxl}
    }
    \label{tab:supp_bitacc_sdxl}
\end{table*}

\begin{table*}[p]
    \centering
    \caption{Bit accuracy comparisons among all adversarial scenarios with GS~\cite{gaussianshading_2024} on SD3.5.}
    \resizebox{1.0\linewidth}{!}{%
        \input{tabular/supp_bitacc_sd35}
    }
    \label{tab:supp_bitacc_sd35}
\end{table*}

\begin{table*}[p]
    \centering
    \caption{Bit accuracy comparisons among all adversarial scenarios with PRCW~\cite{prcwatermark_2025} on SD2.1.}
    \resizebox{1.0\linewidth}{!}{%
        \input{tabular/supp_bitacc_prcw}
    }
    \label{tab:supp_bitacc_prcw}
\end{table*}

\begin{table*}[p]
    \centering
    \caption{Bit accuracy comparisons among all adversarial scenarios with T2SMark~\cite{t2smark_2025} on SD2.1.}
    \resizebox{1.0\linewidth}{!}{%
        \input{tabular/supp_bitacc_t2s}
    }
    \label{tab:supp_bitacc_t2s}
\end{table*}

\FloatBarrier

The average performance over all adversarial settings has already been summarized in \cref{tab:sd21_overview}, \cref{tab:sdxl} and \cref{tab:sd35}  of the main paper. 
Here, we further present per-distortion results to offer a more fine-grained analysis.
Specifically, we report the bit accuracy under each distortion in \cref{tab:supp_bitacc_sd21}, \cref{tab:supp_bitacc_sdxl} and \cref{tab:supp_bitacc_sd35}, corresponding to SD2.1, SDXL and SD3.5 diffusion models, respectively. 
For SD2.1, we additionally provide per-distortion TPR@FPR=1e-6 in \cref{tab:supp_tpr_sd21}. 
These detailed results consistently show that OSI outperforms the GS baseline across all considered distortion types and diffusion backbones.

We also compare OSI with PRCW~\cite{prcwatermark_2025} in \cref{tab:supp_bitacc_prcw}, where we report the per-distortion bit accuracy. 
Although OSI is slightly inferior to PRCW under a few specific distortion settings, it still achieves better overall robustness while requiring only a single inversion step. 
In addition, \cref{tab:supp_bitacc_t2s} presents per-distortion bit accuracy for T2SMark~\cite{t2smark_2025}. 
Across all distortion scenarios, OSI consistently surpasses the original T2SMark.

\begin{figure*}[b]
    \centering
    \includegraphics[width=1\linewidth]{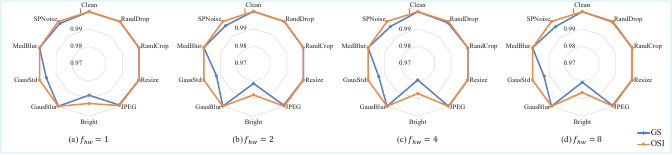}
    \caption{LogAUC values of GS and OSI under different $f_{hw}$ settings across various image distortions.}
    \label{fig:supp_logauc}
\end{figure*}

\begin{table*}[b]
    \centering
    \small
    \caption{Bit accuracy comparisons under advanced attacks on SD2.1.}
    \resizebox{1.0\linewidth}{!}{%
        \input{tabular/supp_advanced_attacks}
    }
    \label{tab:supp_advanced_attacks}
\end{table*}

To obtain an integrated measure that jointly accounts for true positive rate (TPR) and false positive rate (FPR), we further report logAUC. 
Concretely, we scan the FPR on a logarithmic scale from $10^{-12}$ to $10^{-1}$ and compute the area under the resulting ROC curve. 
The visualization in \cref{fig:supp_logauc} shows that our OSI method consistently outperforms GS in logAUC across all $f_{hw}$ configurations and adversarial scenarios, further demonstrating the superiority of OSI.

\subsection{Advanced Attack}
Beyond the standard distortion-based evaluation, we also investigate the robustness of OSI under more advanced attacks.
In particular, we consider two families of attacks: VAE-based compression attacks and embedding-based adversarial attacks.
\cref{tab:supp_advanced_attacks} details the quantitative performance of OSI under these advanced attack scenarios.

The first family consists of attacks based on variational autoencoder (VAE) image compression models. 
Given a watermarked image $I$, the VAE encodes $I$ into a latent representation, applies quantization to induce loss, and then reconstructs the latent back to pixel space. 
The overall compression–reconstruction process can be expressed as
\begin{equation}
    \hat{I}_{q} = \mathcal{D}\!\left(\mathcal{Q}_{q}\!\left(\mathcal{E}(I)\right)\right),
\end{equation}
where $\mathcal{E}$ and $\mathcal{D}$ denote VAE encoder and VAE decoder, and $\mathcal{Q}_q(\cdot)$ is a quantization operator which is controlled by the quality parameter $q$. 
We evaluate six VAE-based compression models~\cite{bmshj2018,mbt2018,cheng2020} at their strongest attack setting (quality level $q=1$). As shown in \cref{tab:supp_advanced_attacks}, OSI consistently outperforms GS under all compression attacks.

The second family, embedding attack, aims to perturb the input image such that its representation under a fixed encoder is significantly altered, while the perturbed image remains visually similar to the original in pixel space.
Let $f(\cdot)$ denote a pretrained embedding model. Given a watermarked image $I$, we first compute its embedding $f(I)$ and then seek an adversarial image $I_{adv}$ that is close to $I$ in pixel space but yields a substantially different embedding $f(I_{adv})$.
Formally, the objective is defined as:
\begin{equation}
    \underset{I_{adv}}{\max}\; \mathbf{Dist}\bigl(f(I_{adv}),\, f(I)\bigr)
    \quad \text{s.t.} \quad \|I_{adv} - I\|_{\infty} \le \delta,
    \label{eq:supp_embedding_attack}
\end{equation}
where $\mathbf{Dist}$ denotes a distance measure, and $\delta$ is the perturbation budget controlling the maximum allowable pixel-wise deviation.
After optimizing \cref{eq:supp_embedding_attack}, $I_{adv}$ is likely to induce failure in watermark extraction.
In our experiment, we instantiate embedding attack using two encoders, ResNet-18~\cite{resnet_2016} and CLIP~\cite{clip_2021}. As shown in \cref{tab:supp_advanced_attacks}, OSI still outperforms GS across all settings.
\begin{figure*}[b]
    \centering
    \includegraphics[width=0.75\linewidth]{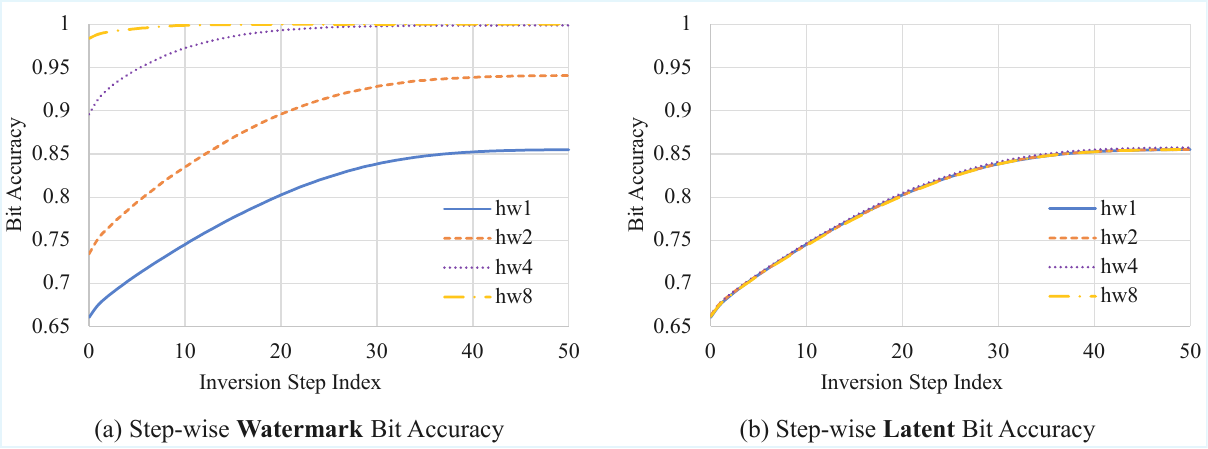}
    \caption{Step-wise Accuracy across different $f_{hw}$ configurations for GS~\cite{gaussianshading_2024}. Watermark bit accuracy is calculated by evaluating the recovery of the watermark ($\frac{c\cdot h\cdot w}{f_{hw}^2}$). Latent bit accuracy is calculated by evaluating the recovery of the initial latent ($c\cdot h\cdot w$).}
    \label{fig:supp_stepacc}
\end{figure*}
\begin{table*}[b]
    \centering
    \small
    \caption{Comparisions among different cryptographic modules for GS.}
    \input{tabular/supp_chacha}
    \label{tab:chacha}
\end{table*}

\section{Additional Studies}
\subsection{Step-wise Bit Accuracy}
\label{subsec:stepacc}
In this part, we present a comprehensive analysis of the step-wise bit accuracy for GS, using latents at each step of the multi-step diffusion inversion process for watermark verification. 
For the 50-step inversion setting, we denote the latent from the encoder output as Step-0 and the latent after the final inversion step as Step-50. 

In \cref{fig:supp_stepacc}(a), we report the watermark ($\frac{c\cdot h\cdot w}{f_{hw}^2}$) bit accuracy in the clean scenario. 
For all $f_{hw}$ configurations, the watermark bit accuracy increases steadily as the inversion step index grows. 
Larger $f_{hw}$ leads to higher accuracy. In particular, even the encoder output Step-0 already achieves over 98\% bit accuracy when $f_{hw}=8$. 
These observations confirm the inherent inversion capability of both the diffusion backbone and the encoder. 

In \cref{fig:supp_stepacc}(b), we directly evaluate the bit accuracy at the latent level ($c\cdot h\cdot w$). 
The resulting curves for different $f_{hw}$ are highly similar, indicating that REP does not change the intrinsic inversion capability of the model.
Instead, REP primarily provides gains at the cryptographic procedure by adding redundancy to the watermark bits.
As discussed in \cref{sec:communication}, the similarity in \cref{fig:supp_stepacc}(b) indicates that different codings ($f_{hw}$) do not change the channel capacity (latent bit accuracy).
Moreover, this similarity also explains why we conduct several ablation studies with a fixed setting $f_{hw}=1$ (see \cref{tab:data aug}, \cref{tab:cfg} and \cref{tab:scheduler}), since varying $f_{hw}$ does not qualitatively affect the underlying inversion behavior.

\subsection{Different Cryptographic Modules}
GS~\cite{gaussianshading_2024} provides two different cryptographic modules, ChaCha20~\cite{chacha_2008} and a simple XOR-based scheme, for encrypting repeated watermarks into sign masks and decrypting them back. 
We conduct an ablation study to compare their impact on overall watermarking performance. 
As shown in \cref{tab:chacha}, the results obtained with them are almost identical, with no meaningful performance gap.
This observation is consistent with our communication system perspective in \cref{sec:communication}, where we disentangle coding from the channel. 
The designs of coding do not affect the underlying channel capacity.

\subsection{Different Total Inversion Steps}
In \cref{subsec:stepacc}, we analyze the performance of the latents at each step of a 50-step diffusion inversion process. 
Here, we further investigate how the total number of inversion steps affects watermark extraction using GS. 
As shown in \cref{tab:supp_invstep}, increasing the total number of inversion steps generally leads to higher bit accuracy. However, the gains quickly saturate, with additional steps yielding only marginal benefits. 
Notably, OSI demonstrates superior performance compared to the computationally intensive 100-step inversion, despite requiring only a single forward step, which further demonstrates both the effectiveness and efficiency of OSI.

\begin{table*}
    \small
    \centering
    \caption{Comparison among different total inversion steps.}
    \input{tabular/supp_invstep}
    \label{tab:supp_invstep}
\end{table*}

\begin{figure}[]
    \centering
    \includegraphics[width=0.8\linewidth]{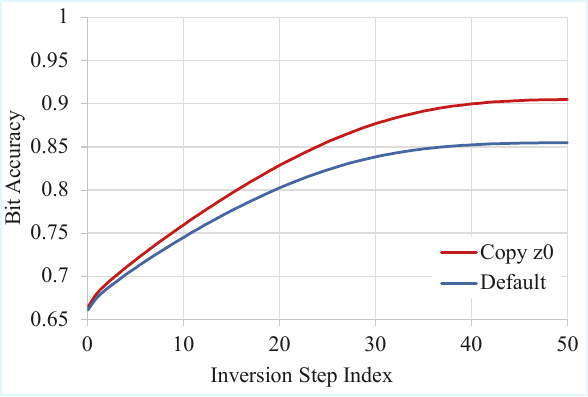}
    \caption{Effect of copying $z_0$ for multi-step diffusion inversion.}
    \label{fig:supp_copyz0}
\end{figure}
\subsection{Alignment of Encoder Output}
In \cref{subsec:OSI}, we introduce $\mathcal{L}_\mathrm{MSE}$ to align the encoder output latent $\hat{z}_0$ during watermark extraction and decoder input latent $z_0$ during image generation. 
This alignment improves the overall watermark extraction performance, as evidenced by the ablation results in \cref{tab:module ablation}. 
Here we provide a complementary experiment based on GS. 
Specifically, during watermark extraction, we directly replace $\hat{z}_0$ with $z_0$ and evaluate the corresponding step-wise bit accuracy. 
As shown in \cref{fig:supp_copyz0}, the bit accuracy increases after the replacement, indicating that better alignment between $\hat{z}_0$ and $z_0$ is crucial for achieving higher extraction accuracy. 
From a communication system perspective, this alignment can be interpreted as a partial improvement of the effective channel, which can be schematically represented as 
\begin{equation}
    wm\to...\to z_0\to \hat{z_0}\to...\to\hat{wm}.
    \label{eq:supp_z0_channel}
\end{equation}
By reducing the noise introduced during the decoding–encoding transition ($z_0\to \hat{z_0}$), the alignment improves the overall channel capacity. Consequently, we explicitly enforce this alignment during finetuning to further improve the effective capacity of our OSI model.

\begin{figure}[]
  \centering
    \includegraphics[width=1\linewidth]{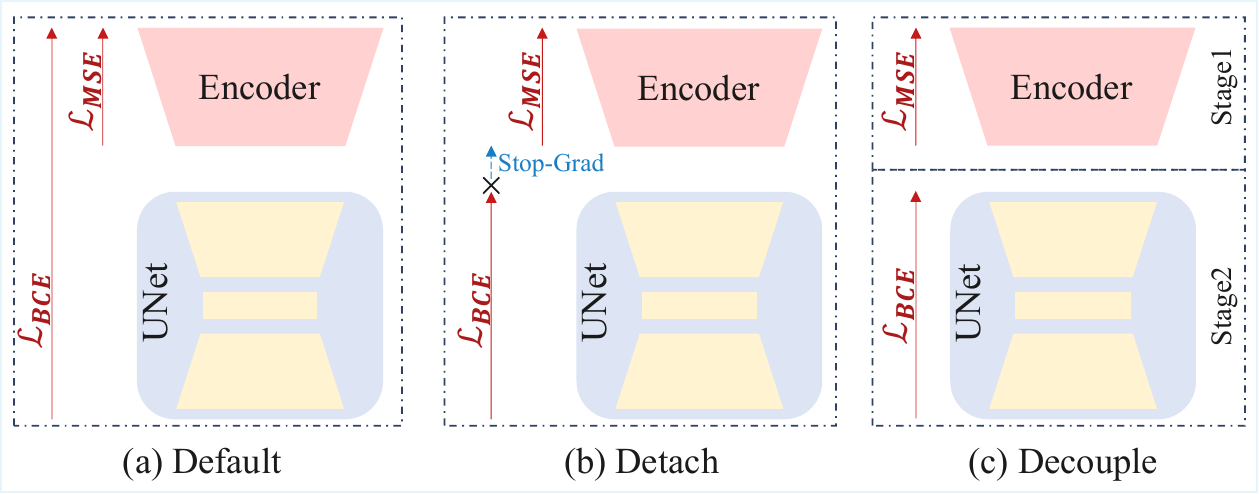}
  \caption{Visualization of 3 finetuning strategies.}
  \label{fig:trainstrategy}
\end{figure}

\begin{table}[]
\small
    \centering
    \caption{Comparison of finetuning strategies.}
    \input{tabular/train_strategy}
    \label{tab:finetunestrategy}
\end{table}

\subsection{Different Finetuning Strategy}
In this section, we describe the finetuning strategies used for OSI.
We introduce $\mathcal{L}_\mathrm{BCE}$ and $\mathcal{L}_\mathrm{MSE}$ to supervise the finetuning of OSI. 
The $\mathcal{L}_\mathrm{MSE}$ term is back-propagated only through the encoder, whereas the $\mathcal{L}_\mathrm{BCE}$ term can be applied under different strategies.
By default, we use $\mathcal{L}_\mathrm{BCE}$ to jointly update both the UNet and the encoder.
In addition, we investigate two variants where only the UNet is updated by $\mathcal{L}_\mathrm{BCE}$. 
The first variant, denoted ``detach'', performs end-to-end finetuning but detaches the gradient between the UNet and the encoder, so that $\mathcal{L}_\mathrm{BCE}$ does not update the encoder parameters.
The second variant, denoted ``decouple'', adopts a two-stage procedure: in stage 1, we finetune the encoder using $\mathcal{L}_\mathrm{MSE}$; in stage 2, we freeze the encoder and finetune only the UNet using $\mathcal{L}_\mathrm{BCE}$.
We illustrate all three strategies in \cref{fig:trainstrategy}.
As summarized in \cref{tab:finetunestrategy}, our default finetuning strategy consistently outperforms both ``detach'' and ``decouple'', and is thus used in all main experiments.

%% file: tabular/supp_tpr_fpr_threshold.tex
\begin{tabular}{ccccc}
    \toprule
    $f_{hw}$ & 1      & 2      & 4      & 8      \\
    \midrule
    $\tau$   & 0.5186 & 0.5371 & 0.5742 & 0.6484 \\
    \bottomrule
\end{tabular}

%% file: tabular/supp_bitacc_sd21.tex
\begin{tabular}{ccccccccccccc}
    \toprule
    Method & $f_{hw}$ & Clean & RandDrop & RandCrop & Resize & JPEG & Bright & GausBlur & GausStd & MedBlur & SPNoise & Adv. Avg \\
    \midrule
    \multirow{4}{*}{GS} & 1 & 0.8560 & 0.6156 & 0.6187 & 0.7050 & 0.6885 & 0.6951 & 0.6516 & 0.6389 & 0.7254 & 0.6151 & 0.6616 \\
     & 2 & 0.9408 & 0.6703 & 0.6761 & 0.7883 & 0.7671 & 0.7634 & 0.7195 & 0.7015 & 0.8135 & 0.6683 & 0.7298 \\
     & 4 & 0.9984 & 0.8223 & 0.8338 & 0.9464 & 0.9200 & 0.8828 & 0.8821 & 0.8462 & 0.9630 & 0.8089 & 0.8784 \\
     & 8 & 1.0000 & 0.9651 & 0.9747 & 0.9972 & 0.9878 & 0.9538 & 0.9849 & 0.9567 & 0.9988 & 0.9365 & 0.9728 \\
     \midrule
    \multirow{4}{*}{\begin{tabular}[c]{@{}c@{}}OSI\\ (ours)\end{tabular}} & 1 & \textbf{0.8839} & \textbf{0.6444} & \textbf{0.6428} & \textbf{0.7576} & \textbf{0.7351} & \textbf{0.7220} & \textbf{0.7227} & \textbf{0.7589} & \textbf{0.7822} & \textbf{0.8622} & \textbf{0.7364} \\
     & 2 & \textbf{0.9607} & \textbf{0.7111} & \textbf{0.7091} & \textbf{0.8501} & \textbf{0.8238} & \textbf{0.7926} & \textbf{0.8105} & \textbf{0.8509} & \textbf{0.8764} & \textbf{0.9453} & \textbf{0.8189} \\
     & 4 & \textbf{0.9996} & \textbf{0.8841} & \textbf{0.8801} & \textbf{0.9815} & \textbf{0.9617} & \textbf{0.9081} & \textbf{0.9632} & \textbf{0.9752} & \textbf{0.9892} & \textbf{0.9986} & \textbf{0.9491} \\
     & 8 & \textbf{1.0000} & \textbf{0.9899} & \textbf{0.9921} & \textbf{0.9997} & \textbf{0.9971} & \textbf{0.9685} & \textbf{0.9992} & \textbf{0.9987} & \textbf{0.9998} & \textbf{1.0000} & \textbf{0.9939} \\
     \bottomrule
\end{tabular}

%% file: tabular/supp_tpr_sd21.tex
\begin{tabular}{ccccccccccccc}
\toprule
Method & $f_{hw}$ & Clean & RandDrop & RandCrop & Resize & JPEG & Bright & GausBlur & GausStd & MedBlur & SPNoise & Adv. Avg \\
\midrule
\multirow{4}{*}{GS} & 1 & 1.0000 & 1.0000 & 1.0000 & 1.0000 & 0.9990 & 0.9890 & 1.0000 & 0.9970 & 1.0000 & 0.9990 & 0.9982 \\
 & 2 & 1.0000 & 1.0000 & 1.0000 & 1.0000 & 0.9990 & 0.9810 & 1.0000 & 0.9920 & 1.0000 & 0.9980 & 0.9967 \\
 & 4 & 1.0000 & 0.9990 & 1.0000 & 1.0000 & 1.0000 & 0.9770 & 1.0000 & 0.9930 & 1.0000 & 0.9990 & 0.9964 \\
 & 8 & 1.0000 & 1.0000 & 1.0000 & 1.0000 & 1.0000 & 0.9810 & 1.0000 & 0.9930 & 1.0000 & 0.9990 & 0.9970 \\
 \midrule
\multirow{4}{*}{\begin{tabular}[c]{@{}c@{}}OSI\\ (ours)\end{tabular}} & 1 & 1.0000 & 1.0000 & 1.0000 & 1.0000 & \textbf{1.0000} & \textbf{0.9920} & 1.0000 & \textbf{1.0000} & 1.0000 & \textbf{1.0000} & \textbf{0.9991} \\
 & 2 & 1.0000 & 1.0000 & 1.0000 & 1.0000 & \textbf{1.0000} & \textbf{0.9860} & 1.0000 & \textbf{1.0000} & 1.0000 & \textbf{1.0000} & \textbf{0.9984} \\
 & 4 & 1.0000 & \textbf{1.0000} & 1.0000 & 1.0000 & 1.0000 & \textbf{0.9870} & 1.0000 & \textbf{1.0000} & 1.0000 & \textbf{1.0000} & \textbf{0.9986} \\
 & 8 & 1.0000 & 1.0000 & 1.0000 & 1.0000 & 1.0000 & \textbf{0.9860} & 1.0000 & \textbf{1.0000} & 1.0000 & \textbf{1.0000} & \textbf{0.9984} \\
 \bottomrule
\end{tabular}

%% file: tabular/supp_bitacc_sdxl.tex
\begin{tabular}{ccccccccccccc}
\toprule
Method & $f_{hw}$ & Clean & RandDrop & RandCrop & Resize & JPEG & Bright & GausBlur & GausStd & MedBlur & SPNoise & Adv. Avg \\
\midrule
\multirow{4}{*}{GS} & 1 & 0.7845 & 0.5935 & 0.5993 & 0.6388 & 0.6195 & 0.6326 & 0.5904 & 0.6098 & 0.6572 & 0.5698 & 0.6123 \\
 & 2 & 0.8789 & 0.6388 & 0.6477 & 0.7027 & 0.6768 & 0.6879 & 0.6340 & 0.6631 & 0.7282 & 0.6049 & 0.6649 \\
 & 4 & 0.9894 & 0.7719 & 0.7878 & 0.8609 & 0.8238 & 0.8016 & 0.7579 & 0.8034 & 0.8918 & 0.7079 & 0.8008 \\
 & 8 & 0.9999 & 0.9267 & 0.9452 & 0.9772 & 0.9523 & 0.8908 & 0.9040 & 0.9400 & 0.9883 & 0.8500 & 0.9305 \\
 \midrule
\multirow{4}{*}{\begin{tabular}[c]{@{}c@{}}OSI\\ (ours)\end{tabular}} & 1 & \textbf{0.8276} & \textbf{0.6154} & \textbf{0.6185} & \textbf{0.6883} & \textbf{0.6595} & \textbf{0.6598} & \textbf{0.6523} & \textbf{0.6717} & \textbf{0.7041} & \textbf{0.7383} & \textbf{0.6676} \\
 & 2 & \textbf{0.9185} & \textbf{0.6708} & \textbf{0.6751} & \textbf{0.7687} & \textbf{0.7319} & \textbf{0.7218} & \textbf{0.7210} & \textbf{0.7479} & \textbf{0.7889} & \textbf{0.8294} & \textbf{0.7395} \\
 & 4 & \textbf{0.9963} & \textbf{0.8237} & \textbf{0.8314} & \textbf{0.9318} & \textbf{0.8936} & \textbf{0.8345} & \textbf{0.8818} & \textbf{0.9096} & \textbf{0.9475} & \textbf{0.9689} & \textbf{0.8914} \\
 & 8 & \textbf{1.0000} & \textbf{0.9646} & \textbf{0.9741} & \textbf{0.9967} & \textbf{0.9864} & \textbf{0.9150} & \textbf{0.9861} & \textbf{0.9907} & \textbf{0.9982} & \textbf{0.9990} & \textbf{0.9790} \\
 \bottomrule
\end{tabular}

%% file: tabular/supp_bitacc_sd35.tex
\begin{tabular}{ccccccccccccc}
\toprule
Method & $f_{hw}$ & Clean & RandDrop & RandCrop & Resize & JPEG & Bright & GausBlur & GausStd & MedBlur & SPNoise & Adv. Avg \\
\midrule
\multirow{4}{*}{GS} & 1 & 0.7604 & 0.5850 & 0.5953 & 0.5800 & 0.5759 & 0.6572 & 0.5541 & 0.5424 & 0.5896 & 0.5499 & 0.5810 \\
 & 2 & 0.8520 & 0.6269 & 0.6424 & 0.6183 & 0.6136 & 0.7244 & 0.5806 & 0.5632 & 0.6330 & 0.5749 & 0.6197 \\
 & 4 & 0.9746 & 0.7489 & 0.7770 & 0.7314 & 0.7220 & 0.8698 & 0.6640 & 0.6270 & 0.7581 & 0.6518 & 0.7278 \\
 & 8 & 0.9982 & 0.8942 & 0.9335 & 0.8722 & 0.8588 & 0.9635 & 0.7919 & 0.7246 & 0.9051 & 0.7713 & 0.8572 \\
 \midrule
\multirow{4}{*}{\begin{tabular}[c]{@{}c@{}}OSI\\ (ours)\end{tabular}} & 1 & \textbf{0.7978} & \textbf{0.5993} & \textbf{0.6087} & \textbf{0.6213} & \textbf{0.6065} & \textbf{0.6818} & \textbf{0.5982} & \textbf{0.6282} & \textbf{0.6434} & \textbf{0.7252} & \textbf{0.6347} \\
 & 2 & \textbf{0.8887} & \textbf{0.6468} & \textbf{0.6611} & \textbf{0.6773} & \textbf{0.6562} & \textbf{0.7520} & \textbf{0.6449} & \textbf{0.6859} & \textbf{0.7082} & \textbf{0.8092} & \textbf{0.6935} \\
 & 4 & \textbf{0.9858} & \textbf{0.7812} & \textbf{0.8068} & \textbf{0.8243} & \textbf{0.7923} & \textbf{0.8940} & \textbf{0.7774} & \textbf{0.8323} & \textbf{0.8666} & \textbf{0.9478} & \textbf{0.8359} \\
 & 8 & \textbf{0.9993} & \textbf{0.9275} & \textbf{0.9569} & \textbf{0.9572} & \textbf{0.9283} & \textbf{0.9732} & \textbf{0.9281} & \textbf{0.9521} & \textbf{0.9786} & \textbf{0.9941} & \textbf{0.9551} \\
 \bottomrule
\end{tabular}

%% file: tabular/supp_bitacc_prcw.tex
\begin{tabular}{ccccccccccccc}
\toprule
Method & Msg. Bit & Clean & RandDrop & RandCrop & Resize & JPEG & Bright & GausBlur & GausStd & MedBlur & SPNoise & Adv. Avg \\
\midrule
\multirow{5}{*}{PRC} & 4096 & 0.8751 & 0.8752 & 0.8751 & 0.8752 & 0.8750 & 0.8750 & \textbf{0.8751} & \textbf{0.8750} & \textbf{0.8748} & \textbf{0.8754} & 0.8751 \\
 & 2048 & 0.9480 & \textbf{0.9376} & 0.9375 & 0.9374 & \textbf{0.9375} & 0.9387 & 0.9374 & 0.9376 & 0.9374 & 0.9374 & 0.9376 \\
 & 1024 & 1.0000 & 0.9688 & 0.9687 & 0.9775 & 0.9778 & 0.9801 & 0.9687 & 0.9696 & 0.9848 & 0.9691 & 0.9739 \\
 & 512 & 1.0000 & 0.9847 & 0.9855 & 0.9975 & 0.9930 & 0.9929 & 0.9872 & 0.9872 & 0.9983 & 0.9860 & 0.9903 \\
 & 256 & 1.0000 & 0.9969 & 0.9980 & 0.9997 & 0.9982 & 0.9970 & 0.9964 & 0.9950 & 0.9997 & 0.9934 & 0.9972 \\
 \midrule
\multirow{5}{*}{\begin{tabular}[c]{@{}c@{}}OSI\\ (ours)\end{tabular}} & 4096 & \textbf{0.8752} & \textbf{0.8753} & 0.8751 & 0.8752 & 0.8750 & \textbf{0.8752} & 0.8750 & 0.8749 & 0.8747 & 0.8752 & 0.8751 \\
 & 2048 & \textbf{0.9605} & 0.9374 & 0.9375 & 0.9374 & 0.9374 & \textbf{0.9399} & \textbf{0.9375} & \textbf{0.9387} & \textbf{0.9377} & \textbf{0.9518} & \textbf{0.9395} \\
 & 1024 & 1.0000 & \textbf{0.9878} & \textbf{0.9878} & \textbf{0.9966} & \textbf{0.9913} & \textbf{0.9856} & \textbf{0.9881} & \textbf{0.9965} & \textbf{0.9984} & \textbf{1.0000} & \textbf{0.9925} \\
 & 512 & 1.0000 & \textbf{1.0000} & \textbf{0.9998} & \textbf{0.9998} & \textbf{0.9994} & \textbf{0.9958} & \textbf{0.9994} & \textbf{0.9995} & \textbf{0.9998} & \textbf{1.0000} & \textbf{0.9993} \\
 & 256 & 1.0000 & \textbf{0.9998} & \textbf{0.9999} & \textbf{1.0000} & \textbf{0.9997} & \textbf{0.9980} & \textbf{0.9999} & \textbf{0.9998} & \textbf{1.0000} & \textbf{1.0000} & \textbf{0.9997} \\
 \bottomrule
\end{tabular}

%% file: tabular/supp_bitacc_t2s.tex
\begin{tabular}{ccccccccccccc}
\toprule
Method & Msg. Bit & Clean & RandDrop & RandCrop & Resize & JPEG & Bright & GausBlur & GausStd & MedBlur & SPNoise & Adv. Avg \\
\midrule
\multirow{5}{*}{T2S} & 4096 & 0.9909 & 0.7204 & 0.7294 & 0.8819 & 0.8429 & 0.8262 & 0.8078 & 0.7562 & 0.9031 & 0.7261 & 0.7993 \\
 & 2048 & 0.9979 & 0.7991 & 0.7784 & 0.9481 & 0.9083 & 0.8776 & 0.8875 & 0.8180 & 0.9612 & 0.7931 & 0.8635 \\
 & 1024 & 0.9998 & 0.9073 & 0.9575 & 0.9863 & 0.9628 & 0.9341 & 0.9542 & 0.8871 & 0.9912 & 0.8677 & 0.9387 \\
 & 512 & 1.0000 & 0.9792 & 0.9915 & 0.9974 & 0.9892 & 0.9661 & 0.9881 & 0.9401 & 0.9987 & 0.9304 & 0.9756 \\
 & 256 & 1.0000 & 0.9984 & 0.9992 & 0.9995 & 0.9967 & 0.9807 & 0.9982 & 0.9748 & 0.9998 & 0.9704 & 0.9909 \\
 \midrule
\multirow{5}{*}{\begin{tabular}[c]{@{}c@{}}OSI\\ (ours)\end{tabular}} & 4096 & \textbf{0.9977} & \textbf{0.7626} & \textbf{0.7642} & \textbf{0.9455} & \textbf{0.9137} & \textbf{0.8616} & \textbf{0.9132} & \textbf{0.9338} & \textbf{0.9611} & \textbf{0.9929} & \textbf{0.8943} \\
 & 2048 & \textbf{0.9998} & \textbf{0.8527} & \textbf{0.8119} & \textbf{0.9863} & \textbf{0.9657} & \textbf{0.9122} & \textbf{0.9713} & \textbf{0.9762} & \textbf{0.9916} & \textbf{0.9986} & \textbf{0.9407} \\
 & 1024 & \textbf{1.0000} & \textbf{0.9595} & \textbf{0.9891} & \textbf{0.9985} & \textbf{0.9936} & \textbf{0.9638} & \textbf{0.9958} & \textbf{0.9963} & \textbf{0.9992} & \textbf{0.9999} & \textbf{0.9884} \\
 & 512 & 1.0000 & \textbf{0.9988} & \textbf{0.9996} & \textbf{0.9999} & \textbf{0.9993} & \textbf{0.9854} & \textbf{0.9997} & \textbf{0.9997} & \textbf{1.0000} & \textbf{1.0000} & \textbf{0.9980} \\
 & 256 & 1.0000 & \textbf{1.0000} & \textbf{1.0000} & \textbf{1.0000} & \textbf{0.9998} & \textbf{0.9920} & \textbf{1.0000} & \textbf{0.9999} & \textbf{1.0000} & \textbf{1.0000} & \textbf{0.9991} \\
 \bottomrule
\end{tabular}

%% file: tabular/supp_advanced_attacks.tex
\begin{tabular}{c c c c c c c c c c }
    \toprule
    \multirow{2}{*}[-0.7ex]{Method} & \multirow{2}{*}[-0.7ex]{$f_{hw}$} &
    \multicolumn{6}{c}{Compression} &
    \multicolumn{2}{c}{Embedding} \\
    \cmidrule(lr){3-8} 
    \cmidrule(lr){9-10}
    & &
    \begin{tabular}{@{}c@{}}bmshj\_factorized\end{tabular} &
    \begin{tabular}{@{}c@{}}bmshj\_hyperprior\end{tabular} &
    \begin{tabular}{@{}c@{}}mbt2018\_mean\end{tabular} &
    mbt2018 &
    cheng\_anchor &
    cheng\_attn &
    CLIP &
    ResNet18 \\
    \midrule
    
    \multirow{4}{*}{GS}
    & 1 & 0.6165 & 0.6222 & 0.6219 & 0.6276 & 0.6314 & 0.6321 & 0.6384 & 0.6403 \\
    & 2 & 0.6707 & 0.6776 & 0.6775 & 0.6853 & 0.6904 & 0.6917 & 0.6992 & 0.7014 \\
    & 4 & 0.8120 & 0.8194 & 0.8204 & 0.8313 & 0.8370 & 0.8381 & 0.8568 & 0.8601 \\
    & 8 & 0.9409 & 0.9406 & 0.9404 & 0.9498 & 0.9524 & 0.9519 & 0.9640 & 0.9653 \\
    \midrule
    
    \multirow{4}{*}{\begin{tabular}[c]{@{}c@{}}OSI\\(ours)\end{tabular}}
    & 1 & \textbf{0.6241} & \textbf{0.6302} & \textbf{0.6310} & \textbf{0.6344} & \textbf{0.6413} & \textbf{0.6423} & \textbf{0.6513} & \textbf{0.6543} \\
    & 2 & \textbf{0.6814} & \textbf{0.6887} & \textbf{0.6901} & \textbf{0.6947} & \textbf{0.7040} & \textbf{0.7058} & \textbf{0.7171} & \textbf{0.7205} \\
    & 4 & \textbf{0.8279} & \textbf{0.8360} & \textbf{0.8375} & \textbf{0.8442} & \textbf{0.8560} & \textbf{0.8586} & \textbf{0.8783} & \textbf{0.8824} \\
    & 8 & \textbf{0.9528} & \textbf{0.9528} & \textbf{0.9550} & \textbf{0.9590} & \textbf{0.9651} & \textbf{0.9656} & \textbf{0.9749} & \textbf{0.9766} \\
    \bottomrule
\end{tabular}

%% file: tabular/supp_chacha.tex
\begin{tabularx}{\textwidth}{X >{\centering\arraybackslash}X>{\centering\arraybackslash}X>{\centering\arraybackslash}X>{\centering\arraybackslash}X>{\centering\arraybackslash}X>{\centering\arraybackslash}X>{\centering\arraybackslash}X>{\centering\arraybackslash}X>{\centering\arraybackslash}X}
\toprule
\multirow{2}{*}{Method} & \multirow{2}{*}{Module} & \multicolumn{4}{c}{Clean} & \multicolumn{4}{c}{Adversarial} \\
\cmidrule(lr){3-6} \cmidrule(lr){7-10}
 &  & 1 & 2 & 4 & 8 & 1 & 2 & 4 & 8 \\
   \midrule
\multirow{2}{*}{GS} & ChaCha20 & 0.8552 & 0.9416 & 0.9984 & 1.0000 & 0.6613 & 0.7286 & 0.8786 & 0.9724 \\
 & XOR & 0.8560 & 0.9408 & 0.9984 & 1.0000 & 0.6616 & 0.7298 & 0.8784 & 0.9728 \\
 \midrule
\multirow{2}{*}{OSI} & ChaCha20 & 0.8839 & 0.9607 & 0.9996 & 1.0000 & 0.7364 & 0.8189 & 0.9491 & 0.9939 \\
 & XOR & 0.8841 & 0.9607 & 0.9996 & 1.0000 & 0.7364 & 0.8179 & 0.9490 & 0.9941 \\
 \bottomrule
\end{tabularx}

%% file: tabular/supp_invstep.tex
\begin{tabularx}{\textwidth}{X >{\centering\arraybackslash}X>{\centering\arraybackslash}X>{\centering\arraybackslash}X>{\centering\arraybackslash}X>{\centering\arraybackslash}X>{\centering\arraybackslash}X>{\centering\arraybackslash}X>{\centering\arraybackslash}X}
\toprule
 & \multicolumn{7}{c}{GS} & OSI \\ \cmidrule(lr){2-8} \cmidrule(lr){9-9}
 & 1 & 10 & 20 & 30 & 40 & 50 & 100 & 1 \\ \midrule
Clean & 0.7241 & 0.8260 & 0.8433 & 0.8500 & 0.8538 & 0.8560 & 0.8604 & \textbf{0.8839} \\ 
Adversarial & 0.5968 & 0.6116 & 0.6139 & 0.6146 & 0.6149 & 0.6151 & 0.6155 & \textbf{0.7364} \\
\bottomrule
\end{tabularx}

%% file: tabular/train_strategy.tex
\begin{tabularx}{\linewidth}{>{\centering\arraybackslash}X>{\centering\arraybackslash}X>{\centering\arraybackslash}X>{\centering\arraybackslash}X}
    \toprule
     & Default & Detach & Decouple \\ \midrule
    Clean & \textbf{0.8839} & 0.8714 & 0.8725 \\
    Adversarial & \textbf{0.7364} & 0.7227 & 0.7231 \\ \bottomrule
\end{tabularx}

%% file: main.bib
@String(CVPR= {IEEE Conf. Comput. Vis. Pattern Recog.})

@String(ICCV= {Int. Conf. Comput. Vis.})

@String(ECCV= {Eur. Conf. Comput. Vis.})

@String(ICLR = {Int. Conf. Learn. Represent.})

@String(IJCAI = {IJCAI})

@String(CVPR  = {CVPR})

@String(ICCV  = {ICCV})

@String(ECCV  = {ECCV})

@String(ICLR  = {ICLR})

@article{dct_1998,
  author       = {Mauro Barni and
                  Franco Bartolini and
                  Vito Cappellini and
                  Alessandro Piva},
  title        = {A DCT-domain system for robust image watermarking},
  journal      = {Signal Process.},
  volume       = {66},
  number       = {3},
  pages        = {357--372},
  year         = {1998},
  url          = {https://doi.org/10.1016/S0165-1684(98)00015-2},
  doi          = {10.1016/S0165-1684(98)00015-2},
  bibsource    = {dblp computer science bibliography, https://dblp.org}
}

@article{qim_2001,
  author       = {Brian Chen and
                  Gregory W. Wornell},
  title        = {Quantization index modulation: {A} class of provably good methods
                  for digital watermarking and information embedding},
  journal      = {{IEEE} Trans. Inf. Theory},
  volume       = {47},
  number       = {4},
  pages        = {1423--1443},
  year         = {2001},
  url          = {https://doi.org/10.1109/18.923725},
  doi          = {10.1109/18.923725},
  bibsource    = {dblp computer science bibliography, https://dblp.org}
}

@inproceedings{stablesignature_2023,
  author       = {Pierre Fernandez and
                  Guillaume Couairon and
                  Herv{\'{e}} J{\'{e}}gou and
                  Matthijs Douze and
                  Teddy Furon},
  title        = {The Stable Signature: Rooting Watermarks in Latent Diffusion Models},
  booktitle    = {{IEEE/CVF} International Conference on Computer Vision, {ICCV} 2023,
                  Paris, France, October 1-6, 2023},
  pages        = {22409--22420},
  publisher    = {{IEEE}},
  year         = {2023},
  url          = {https://doi.org/10.1109/ICCV51070.2023.02053},
  doi          = {10.1109/ICCV51070.2023.02053},
  bibsource    = {dblp computer science bibliography, https://dblp.org}
}

@inproceedings{aqualora_2024,
  author       = {Weitao Feng and
                  Wenbo Zhou and
                  Jiyan He and
                  Jie Zhang and
                  Tianyi Wei and
                  Guanlin Li and
                  Tianwei Zhang and
                  Weiming Zhang and
                  Nenghai Yu},
  title        = {AquaLoRA: Toward White-box Protection for Customized Stable Diffusion
                  Models via Watermark LoRA},
  booktitle    = {Forty-first International Conference on Machine Learning, {ICML} 2024,
                  Vienna, Austria, July 21-27, 2024},
  publisher    = {OpenReview.net},
  year         = {2024},
  url          = {https://openreview.net/forum?id=8xKGZsnV2a},
  bibsource    = {dblp computer science bibliography, https://dblp.org}
}

@inproceedings{gaussianshading_2024,
  author       = {Zijin Yang and
                  Kai Zeng and
                  Kejiang Chen and
                  Han Fang and
                  Weiming Zhang and
                  Nenghai Yu},
  title        = {Gaussian Shading: Provable Performance-Lossless Image Watermarking
                  for Diffusion Models},
  booktitle    = {{IEEE/CVF} Conference on Computer Vision and Pattern Recognition,
                  {CVPR} 2024, Seattle, WA, USA, June 16-22, 2024},
  pages        = {12162--12171},
  publisher    = {{IEEE}},
  year         = {2024},
  url          = {https://doi.org/10.1109/CVPR52733.2024.01156},
  doi          = {10.1109/CVPR52733.2024.01156},
  bibsource    = {dblp computer science bibliography, https://dblp.org}
}

@inproceedings{treering_2023,
  author       = {Yuxin Wen and
                  John Kirchenbauer and
                  Jonas Geiping and
                  Tom Goldstein},
  editor       = {Alice Oh and
                  Tristan Naumann and
                  Amir Globerson and
                  Kate Saenko and
                  Moritz Hardt and
                  Sergey Levine},
  title        = {Tree-Rings Watermarks: Invisible Fingerprints for Diffusion Images},
  booktitle    = {Advances in Neural Information Processing Systems 36: Annual Conference
                  on Neural Information Processing Systems 2023, NeurIPS 2023, New Orleans,
                  LA, USA, December 10 - 16, 2023},
  year         = {2023},
  url          = {http://papers.nips.cc/paper\_files/paper/2023/hash/b54d1757c190ba20dbc4f9e4a2f54149-Abstract-Conference.html},
  bibsource    = {dblp computer science bibliography, https://dblp.org}
}

@inproceedings{prcwatermark_2025,
  author       = {Sam Gunn and
                  Xuandong Zhao and
                  Dawn Song},
  title        = {An Undetectable Watermark for Generative Image Models},
  booktitle    = {The Thirteenth International Conference on Learning Representations,
                  {ICLR} 2025, Singapore, April 24-28, 2025},
  publisher    = {OpenReview.net},
  year         = {2025},
  url          = {https://openreview.net/forum?id=jlhBFm7T2J},
  bibsource    = {dblp computer science bibliography, https://dblp.org}
}

@inproceedings{gaussmarker_2025,
  title     = {GaussMarker: Robust Dual-Domain Watermark for Diffusion Models},
  author    = {Li, Kecen and Huang, Zhicong and Hou, Xinwen and Hong, Cheng},
  booktitle = {Proceedings of the 42nd International Conference on Machine Learning},
  pages     = {34688--34701},
  year      = {2025},
  editor    = {Singh, Aarti and Fazel, Maryam and Hsu, Daniel and Lacoste-Julien, Simon and Berkenkamp, Felix and Maharaj, Tegan and Wagstaff, Kiri and Zhu, Jerry},
  volume    = {267},
  series    = {Proceedings of Machine Learning Research},
  month     = {13--19 Jul},
  publisher = {PMLR},
  url       = {https://proceedings.mlr.press/v267/li25ae.html}
}

@InProceedings{tagwm_2025,
    author    = {Chen, Yuzhuo and Ma, Zehua and Fang, Han and Zhang, Weiming and Yu, Nenghai},
    title     = {TAG-WM: Tamper-Aware Generative Image Watermarking via Diffusion Inversion Sensitivity},
    booktitle = {Proceedings of the IEEE/CVF International Conference on Computer Vision (ICCV)},
    month     = {October},
    year      = {2025},
    pages     = {16723-16732}
}

@inproceedings{maxsive_2025,
  author    = {Mao, Po-Yuan and Tsai, Cheng-Chang and Lu, Chun-Shien},
  title     = {MaXsive: High-Capacity and Robust Training-Free Generative Image Watermarking in Diffusion Models},
  booktitle = {Proceedings of the 33rd ACM International Conference on Multimedia},
  series    = {MM '25},
  year      = {2025},
  location  = {Dublin, Ireland},
  publisher = {Association for Computing Machinery},
  address   = {New York, NY, USA},
  isbn      = {979-8-4007-2035-2/2025/10},
  doi       = {10.1145/3746027.3755266},
  url       = {https://doi.org/10.1145/3746027.3755266}
}

@inproceedings{secureandefficient_2025,
  author       = {Liangqi Lei and
                  Keke Gai and
                  Jing Yu and
                  Liehuang Zhu and
                  Qi Wu},
  title        = {Secure and Efficient Watermarking for Latent Diffusion Models in Model
                  Distribution Scenarios},
  booktitle    = {Proceedings of the Thirty-Fourth International Joint Conference on
                  Artificial Intelligence, {IJCAI} 2025, Montreal, Canada, August 16-22,
                  2025},
  pages        = {7473--7481},
  publisher    = {ijcai.org},
  year         = {2025},
  url          = {https://doi.org/10.24963/ijcai.2025/831},
  doi          = {10.24963/IJCAI.2025/831},
  bibsource    = {dblp computer science bibliography, https://dblp.org}
}

@inproceedings{sfwmark_2025,
  title={Semantic Watermarking Reinvented: Enhancing Robustness and Generation Quality with Fourier Integrity},
  author={Lee, Sung Ju and Cho, Nam Ik},
  booktitle={Proceedings of the IEEE/CVF International Conference on Computer Vision},
  pages={18759--18769},
  year={2025}
}

@inproceedings{ringid_2024,
  author       = {Hai Ci and
                  Pei Yang and
                  Yiren Song and
                  Mike Zheng Shou},
  editor       = {Ales Leonardis and
                  Elisa Ricci and
                  Stefan Roth and
                  Olga Russakovsky and
                  Torsten Sattler and
                  G{\"{u}}l Varol},
  title        = {RingID: Rethinking Tree-Ring Watermarking for Enhanced Multi-key Identification},
  booktitle    = {Computer Vision - {ECCV} 2024 - 18th European Conference, Milan, Italy,
                  September 29-October 4, 2024, Proceedings, Part {XXVIII}},
  series       = {Lecture Notes in Computer Science},
  volume       = {15086},
  pages        = {338--354},
  publisher    = {Springer},
  year         = {2024},
  url          = {https://doi.org/10.1007/978-3-031-73390-1\_20},
  doi          = {10.1007/978-3-031-73390-1\_20},
  bibsource    = {dblp computer science bibliography, https://dblp.org}
}

@misc{t2smark_2025,
      title={T2SMark: Balancing Robustness and Diversity in Noise-as-Watermark for Diffusion Models}, 
      author={Jindong Yang and Han Fang and Weiming Zhang and Nenghai Yu and Kejiang Chen},
      year={2025},
      eprint={2510.22366},
      archivePrefix={arXiv},
      primaryClass={cs.CV},
      url={https://arxiv.org/abs/2510.22366}, 
}

@misc{gaussianshading++_2025,
      title={Gaussian Shading++: Rethinking the Realistic Deployment Challenge of Performance-Lossless Image Watermark for Diffusion Models}, 
      author={Zijin Yang and Xin Zhang and Kejiang Chen and Kai Zeng and Qiyi Yao and Han Fang and Weiming Zhang and Nenghai Yu},
      year={2025},
      eprint={2504.15026},
      archivePrefix={arXiv},
      primaryClass={cs.CV},
      url={https://arxiv.org/abs/2504.15026}, 
}

@inproceedings{ddim_2021,
  author       = {Jiaming Song and
                  Chenlin Meng and
                  Stefano Ermon},
  title        = {Denoising Diffusion Implicit Models},
  booktitle    = {9th International Conference on Learning Representations, {ICLR} 2021,
                  Virtual Event, Austria, May 3-7, 2021},
  publisher    = {OpenReview.net},
  year         = {2021},
  url          = {https://openreview.net/forum?id=St1giarCHLP},
  bibsource    = {dblp computer science bibliography, https://dblp.org}
}

@inproceedings{flowstraight_2023,
  author       = {Xingchao Liu and
                  Chengyue Gong and
                  Qiang Liu},
  title        = {Flow Straight and Fast: Learning to Generate and Transfer Data with
                  Rectified Flow},
  booktitle    = {The Eleventh International Conference on Learning Representations,
                  {ICLR} 2023, Kigali, Rwanda, May 1-5, 2023},
  publisher    = {OpenReview.net},
  year         = {2023},
  url          = {https://openreview.net/forum?id=XVjTT1nw5z},
  bibsource    = {dblp computer science bibliography, https://dblp.org}
}

@inproceedings{flowmatching_2023,
  author       = {Yaron Lipman and
                  Ricky T. Q. Chen and
                  Heli Ben{-}Hamu and
                  Maximilian Nickel and
                  Matthew Le},
  title        = {Flow Matching for Generative Modeling},
  booktitle    = {The Eleventh International Conference on Learning Representations,
                  {ICLR} 2023, Kigali, Rwanda, May 1-5, 2023},
  publisher    = {OpenReview.net},
  year         = {2023},
  url          = {https://openreview.net/forum?id=PqvMRDCJT9t},
  bibsource    = {dblp computer science bibliography, https://dblp.org}
}

@inproceedings{dit_2023,
  author       = {William Peebles and
                  Saining Xie},
  title        = {Scalable Diffusion Models with Transformers},
  booktitle    = {{IEEE/CVF} International Conference on Computer Vision, {ICCV} 2023,
                  Paris, France, October 1-6, 2023},
  pages        = {4172--4182},
  publisher    = {{IEEE}},
  year         = {2023},
  url          = {https://doi.org/10.1109/ICCV51070.2023.00387},
  doi          = {10.1109/ICCV51070.2023.00387},
  bibsource    = {dblp computer science bibliography, https://dblp.org}
}

@inproceedings{robin_2024,
  author       = {Huayang Huang and
                  Yu Wu and
                  Qian Wang},
  editor       = {Amir Globersons and
                  Lester Mackey and
                  Danielle Belgrave and
                  Angela Fan and
                  Ulrich Paquet and
                  Jakub M. Tomczak and
                  Cheng Zhang},
  title        = {{ROBIN:} Robust and Invisible Watermarks for Diffusion Models with
                  Adversarial Optimization},
  booktitle    = {Advances in Neural Information Processing Systems 38: Annual Conference
                  on Neural Information Processing Systems 2024, NeurIPS 2024, Vancouver,
                  BC, Canada, December 10 - 15, 2024},
  year         = {2024},
  url          = {http://papers.nips.cc/paper\_files/paper/2024/hash/073c8584ef86bee26fe9d639ec648e28-Abstract-Conference.html},
  bibsource    = {dblp computer science bibliography, https://dblp.org}
}

@inproceedings{
swaldm_2025,
title={{SWA}-{LDM}: Toward Stealthy Watermarks for Latent Diffusion Models},
author={Zhonghao Yang and Linye Lyu and Xuanhang Chang and Daojing He and YU LI},
booktitle={The 1st Workshop on GenAI Watermarking},
year={2025},
url={https://openreview.net/forum?id=wXrrxqQGzq}
}

@techreport{chacha_2008,
  author = {Bernstein, Daniel J.},
  title  = {ChaCha, a variant of Salsa20},
  year   = {2008},
  url    = {https://cr.yp.to/chacha/chacha-20080128.pdf}
}

@article{prcode_2024,
  author       = {Miranda Christ and
                  Sam Gunn},
  title        = {Pseudorandom Error-Correcting Codes},
  journal      = {{IACR} Cryptol. ePrint Arch.},
  pages        = {235},
  year         = {2024},
  url          = {https://eprint.iacr.org/2024/235},
  bibsource    = {dblp computer science bibliography, https://dblp.org}
}

@inproceedings{zodiac_2024,
  author       = {Lijun Zhang and
                  Xiao Liu and
                  Antoni Viros Martin and
                  Cindy Xiong Bearfield and
                  Yuriy Brun and
                  Hui Guan},
  editor       = {Amir Globersons and
                  Lester Mackey and
                  Danielle Belgrave and
                  Angela Fan and
                  Ulrich Paquet and
                  Jakub M. Tomczak and
                  Cheng Zhang},
  title        = {Attack-Resilient Image Watermarking Using Stable Diffusion},
  booktitle    = {Advances in Neural Information Processing Systems 38: Annual Conference
                  on Neural Information Processing Systems 2024, NeurIPS 2024, Vancouver,
                  BC, Canada, December 10 - 15, 2024},
  year         = {2024},
  url          = {http://papers.nips.cc/paper\_files/paper/2024/hash/43d33182360378d5c8e69dd706c24f2f-Abstract-Conference.html},
  bibsource    = {dblp computer science bibliography, https://dblp.org}
}

@article{watermarkdm_2023,
  author       = {Yunqing Zhao and
                  Tianyu Pang and
                  Chao Du and
                  Xiao Yang and
                  Ngai{-}Man Cheung and
                  Min Lin},
  title        = {A Recipe for Watermarking Diffusion Models},
  journal      = {CoRR},
  volume       = {abs/2303.10137},
  year         = {2023},
  url          = {https://doi.org/10.48550/arXiv.2303.10137},
  doi          = {10.48550/ARXIV.2303.10137},
  eprinttype    = {arXiv},
  eprint       = {2303.10137},
  bibsource    = {dblp computer science bibliography, https://dblp.org}
}

@inproceedings{lora_2022,
  author       = {Edward J. Hu and
                  Yelong Shen and
                  Phillip Wallis and
                  Zeyuan Allen{-}Zhu and
                  Yuanzhi Li and
                  Shean Wang and
                  Lu Wang and
                  Weizhu Chen},
  title        = {LoRA: Low-Rank Adaptation of Large Language Models},
  booktitle    = {The Tenth International Conference on Learning Representations, {ICLR}
                  2022, Virtual Event, April 25-29, 2022},
  publisher    = {OpenReview.net},
  year         = {2022},
  url          = {https://openreview.net/forum?id=nZeVKeeFYf9},
  bibsource    = {dblp computer science bibliography, https://dblp.org}
}

@inproceedings{ldm_2022,
  author       = {Robin Rombach and
                  Andreas Blattmann and
                  Dominik Lorenz and
                  Patrick Esser and
                  Bj{\"{o}}rn Ommer},
  title        = {High-Resolution Image Synthesis with Latent Diffusion Models},
  booktitle    = {{IEEE/CVF} Conference on Computer Vision and Pattern Recognition,
                  {CVPR} 2022, New Orleans, LA, USA, June 18-24, 2022},
  pages        = {10674--10685},
  publisher    = {{IEEE}},
  year         = {2022},
  url          = {https://doi.org/10.1109/CVPR52688.2022.01042},
  doi          = {10.1109/CVPR52688.2022.01042},
  bibsource    = {dblp computer science bibliography, https://dblp.org}
}

@inproceedings{sdxl_2024,
  author       = {Dustin Podell and
                  Zion English and
                  Kyle Lacey and
                  Andreas Blattmann and
                  Tim Dockhorn and
                  Jonas M{\"{u}}ller and
                  Joe Penna and
                  Robin Rombach},
  title        = {{SDXL:} Improving Latent Diffusion Models for High-Resolution Image
                  Synthesis},
  booktitle    = {The Twelfth International Conference on Learning Representations,
                  {ICLR} 2024, Vienna, Austria, May 7-11, 2024},
  publisher    = {OpenReview.net},
  year         = {2024},
  url          = {https://openreview.net/forum?id=di52zR8xgf},
  bibsource    = {dblp computer science bibliography, https://dblp.org}
}

@inproceedings{sd3_2024,
  author       = {Patrick Esser and
                  Sumith Kulal and
                  Andreas Blattmann and
                  Rahim Entezari and
                  Jonas M{\"{u}}ller and
                  Harry Saini and
                  Yam Levi and
                  Dominik Lorenz and
                  Axel Sauer and
                  Frederic Boesel and
                  Dustin Podell and
                  Tim Dockhorn and
                  Zion English and
                  Robin Rombach},
  title        = {Scaling Rectified Flow Transformers for High-Resolution Image Synthesis},
  booktitle    = {Forty-first International Conference on Machine Learning, {ICML} 2024,
                  Vienna, Austria, July 21-27, 2024},
  publisher    = {OpenReview.net},
  year         = {2024},
  url          = {https://openreview.net/forum?id=FPnUhsQJ5B},
  bibsource    = {dblp computer science bibliography, https://dblp.org}
}

@inproceedings{dpmsolver_2022,
  author       = {Cheng Lu and
                  Yuhao Zhou and
                  Fan Bao and
                  Jianfei Chen and
                  Chongxuan Li and
                  Jun Zhu},
  editor       = {Sanmi Koyejo and
                  S. Mohamed and
                  A. Agarwal and
                  Danielle Belgrave and
                  K. Cho and
                  A. Oh},
  title        = {DPM-Solver: {A} Fast {ODE} Solver for Diffusion Probabilistic Model
                  Sampling in Around 10 Steps},
  booktitle    = {Advances in Neural Information Processing Systems 35: Annual Conference
                  on Neural Information Processing Systems 2022, NeurIPS 2022, New Orleans,
                  LA, USA, November 28 - December 9, 2022},
  year         = {2022},
  url          = {http://papers.nips.cc/paper\_files/paper/2022/hash/260a14acce2a89dad36adc8eefe7c59e-Abstract-Conference.html},
  bibsource    = {dblp computer science bibliography, https://dblp.org}
}

@inproceedings{mscoco_2014,
  author       = {Tsung{-}Yi Lin and
                  Michael Maire and
                  Serge J. Belongie and
                  James Hays and
                  Pietro Perona and
                  Deva Ramanan and
                  Piotr Doll{\'{a}}r and
                  C. Lawrence Zitnick},
  editor       = {David J. Fleet and
                  Tom{\'{a}}s Pajdla and
                  Bernt Schiele and
                  Tinne Tuytelaars},
  title        = {Microsoft {COCO:} Common Objects in Context},
  booktitle    = {Computer Vision - {ECCV} 2014 - 13th European Conference, Zurich,
                  Switzerland, September 6-12, 2014, Proceedings, Part {V}},
  series       = {Lecture Notes in Computer Science},
  volume       = {8693},
  pages        = {740--755},
  publisher    = {Springer},
  year         = {2014},
  url          = {https://doi.org/10.1007/978-3-319-10602-1\_48},
  doi          = {10.1007/978-3-319-10602-1\_48},
  bibsource    = {dblp computer science bibliography, https://dblp.org}
}

@misc{sdp_2024,
  author       = {Gustavosta},
  title        = {Stable-Diffusion-Prompts},
  year         = {2024},
  howpublished = {\url{https://huggingface.co/datasets/Gustavosta/Stable-Diffusion-Prompts}},
  note         = {Accessed: 2025-11-01}
}

@inproceedings{adam_2015,
  author       = {Diederik P. Kingma and
                  Jimmy Ba},
  editor       = {Yoshua Bengio and
                  Yann LeCun},
  title        = {Adam: A Method for Stochastic Optimization},
  booktitle    = {3rd International Conference on Learning Representations, {ICLR} 2015,
                  San Diego, CA, USA, May 7-9, 2015, Conference Track Proceedings},
  year         = {2015},
  url          = {http://arxiv.org/abs/1412.6980},
  bibsource    = {dblp computer science bibliography, https://dblp.org}
}

@inproceedings{fid_2017,
  author       = {Martin Heusel and
                  Hubert Ramsauer and
                  Thomas Unterthiner and
                  Bernhard Nessler and
                  Sepp Hochreiter},
  editor       = {Isabelle Guyon and
                  Ulrike von Luxburg and
                  Samy Bengio and
                  Hanna M. Wallach and
                  Rob Fergus and
                  S. V. N. Vishwanathan and
                  Roman Garnett},
  title        = {GANs Trained by a Two Time-Scale Update Rule Converge to a Local Nash
                  Equilibrium},
  booktitle    = {Advances in Neural Information Processing Systems 30: Annual Conference
                  on Neural Information Processing Systems 2017, December 4-9, 2017,
                  Long Beach, CA, {USA}},
  pages        = {6626--6637},
  year         = {2017},
  url          = {https://proceedings.neurips.cc/paper/2017/hash/8a1d694707eb0fefe65871369074926d-Abstract.html},
  bibsource    = {dblp computer science bibliography, https://dblp.org}
}

@inproceedings{clip_2021,
  author       = {Alec Radford and
                  Jong Wook Kim and
                  Chris Hallacy and
                  Aditya Ramesh and
                  Gabriel Goh and
                  Sandhini Agarwal and
                  Girish Sastry and
                  Amanda Askell and
                  Pamela Mishkin and
                  Jack Clark and
                  Gretchen Krueger and
                  Ilya Sutskever},
  editor       = {Marina Meila and
                  Tong Zhang},
  title        = {Learning Transferable Visual Models From Natural Language Supervision},
  booktitle    = {Proceedings of the 38th International Conference on Machine Learning,
                  {ICML} 2021, 18-24 July 2021, Virtual Event},
  series       = {Proceedings of Machine Learning Research},
  volume       = {139},
  pages        = {8748--8763},
  publisher    = {{PMLR}},
  year         = {2021},
  url          = {http://proceedings.mlr.press/v139/radford21a.html},
  bibsource    = {dblp computer science bibliography, https://dblp.org}
}

@inproceedings{onestepdiffusion_2024,
  author       = {Weijian Luo and
                  Zemin Huang and
                  Zhengyang Geng and
                  J. Zico Kolter and
                  Guo{-}Jun Qi},
  editor       = {Amir Globersons and
                  Lester Mackey and
                  Danielle Belgrave and
                  Angela Fan and
                  Ulrich Paquet and
                  Jakub M. Tomczak and
                  Cheng Zhang},
  title        = {One-Step Diffusion Distillation through Score Implicit Matching},
  booktitle    = {Advances in Neural Information Processing Systems 38: Annual Conference
                  on Neural Information Processing Systems 2024, NeurIPS 2024, Vancouver,
                  BC, Canada, December 10 - 15, 2024},
  year         = {2024},
  url          = {http://papers.nips.cc/paper\_files/paper/2024/hash/d107ca794d83c8242e357e6a43a068f4-Abstract-Conference.html},
  bibsource    = {dblp computer science bibliography, https://dblp.org}
}

@inproceedings{onestepshortcut_2025,
  author       = {Kevin Frans and
                  Danijar Hafner and
                  Sergey Levine and
                  Pieter Abbeel},
  title        = {One Step Diffusion via Shortcut Models},
  booktitle    = {The Thirteenth International Conference on Learning Representations,
                  {ICLR} 2025, Singapore, April 24-28, 2025},
  publisher    = {OpenReview.net},
  year         = {2025},
  url          = {https://openreview.net/forum?id=OlzB6LnXcS},
  bibsource    = {dblp computer science bibliography, https://dblp.org}
}

@misc{meanflow_2025,
      title={Mean Flows for One-step Generative Modeling}, 
      author={Zhengyang Geng and Mingyang Deng and Xingjian Bai and J. Zico Kolter and Kaiming He},
      year={2025},
      eprint={2505.13447},
      archivePrefix={arXiv},
      primaryClass={cs.LG},
      url={https://arxiv.org/abs/2505.13447}, 
}

@article{cfg_2022,
  author       = {Jonathan Ho and
                  Tim Salimans},
  title        = {Classifier-Free Diffusion Guidance},
  journal      = {CoRR},
  volume       = {abs/2207.12598},
  year         = {2022},
  url          = {https://doi.org/10.48550/arXiv.2207.12598},
  doi          = {10.48550/ARXIV.2207.12598},
  eprinttype    = {arXiv},
  eprint       = {2207.12598},
  bibsource    = {dblp computer science bibliography, https://dblp.org}
}

@inproceedings{edict_2023,
  author       = {Bram Wallace and
                  Akash Gokul and
                  Nikhil Naik},
  title        = {{EDICT:} Exact Diffusion Inversion via Coupled Transformations},
  booktitle    = {{IEEE/CVF} Conference on Computer Vision and Pattern Recognition,
                  {CVPR} 2023, Vancouver, BC, Canada, June 17-24, 2023},
  pages        = {22532--22541},
  publisher    = {{IEEE}},
  year         = {2023},
  url          = {https://doi.org/10.1109/CVPR52729.2023.02158},
  doi          = {10.1109/CVPR52729.2023.02158},
  bibsource    = {dblp computer science bibliography, https://dblp.org}
}

@inproceedings{onexact_2024,
  author       = {Seongmin Hong and
                  Kyeonghyun Lee and
                  Suh Yoon Jeon and
                  Hyewon Bae and
                  Se Young Chun},
  title        = {On Exact Inversion of DPM-Solvers},
  booktitle    = {{IEEE/CVF} Conference on Computer Vision and Pattern Recognition,
                  {CVPR} 2024, Seattle, WA, USA, June 16-22, 2024},
  pages        = {7069--7078},
  publisher    = {{IEEE}},
  year         = {2024},
  url          = {https://doi.org/10.1109/CVPR52733.2024.00675},
  doi          = {10.1109/CVPR52733.2024.00675},
  bibsource    = {dblp computer science bibliography, https://dblp.org}
}

@misc{klingai_website,
  author       = {Kuaishou Technology},
  title        = {Kling AI: AI Image and Video Generation Platform},
  howpublished = {\url{https://klingai.com}},
  note         = {Accessed: Nov. 5, 2025},
  year         = {2025}
}

@misc{midjourney_website,
  author       = {Midjourney, Inc.},
  title        = {Midjourney: AI Image Generation Platform},
  howpublished = {\url{https://www.midjourney.com/}},
  note         = {Accessed: Nov. 5, 2025},
  year         = {2025}
}

@misc{runway_website,
  author       = {Runway AI, Inc.},
  title        = {Runway: AI Image and Video Generator},
  howpublished = {\url{https://runwayml.com/}},
  note         = {Accessed: Nov. 5, 2025},
  year         = {2025}
}

@inproceedings{gradientfreedecoderinversion_2024,
  author       = {Seongmin Hong and
                  Suh Yoon Jeon and
                  Kyeonghyun Lee and
                  Ernest K. Ryu and
                  Se Young Chun},
  editor       = {Amir Globersons and
                  Lester Mackey and
                  Danielle Belgrave and
                  Angela Fan and
                  Ulrich Paquet and
                  Jakub M. Tomczak and
                  Cheng Zhang},
  title        = {Gradient-free Decoder Inversion in Latent Diffusion Models},
  booktitle    = {Advances in Neural Information Processing Systems 38: Annual Conference
                  on Neural Information Processing Systems 2024, NeurIPS 2024, Vancouver,
                  BC, Canada, December 10 - 15, 2024},
  year         = {2024},
  url          = {http://papers.nips.cc/paper\_files/paper/2024/hash/970f59b22f4c72aec75174aae63c7459-Abstract-Conference.html},
  bibsource    = {dblp computer science bibliography, https://dblp.org}
}

@inproceedings{bmshj2018,
  author       = {Johannes Ball{\'{e}} and
                  David Minnen and
                  Saurabh Singh and
                  Sung Jin Hwang and
                  Nick Johnston},
  title        = {Variational image compression with a scale hyperprior},
  booktitle    = {6th International Conference on Learning Representations, {ICLR} 2018,
                  Vancouver, BC, Canada, April 30 - May 3, 2018, Conference Track Proceedings},
  publisher    = {OpenReview.net},
  year         = {2018},
  url          = {https://openreview.net/forum?id=rkcQFMZRb},
  bibsource    = {dblp computer science bibliography, https://dblp.org}
}

@inproceedings{mbt2018,
  author       = {David Minnen and
                  Johannes Ball{\'{e}} and
                  George Toderici},
  editor       = {Samy Bengio and
                  Hanna M. Wallach and
                  Hugo Larochelle and
                  Kristen Grauman and
                  Nicol{\`{o}} Cesa{-}Bianchi and
                  Roman Garnett},
  title        = {Joint Autoregressive and Hierarchical Priors for Learned Image Compression},
  booktitle    = {Advances in Neural Information Processing Systems 31: Annual Conference
                  on Neural Information Processing Systems 2018, NeurIPS 2018, December
                  3-8, 2018, Montr{\'{e}}al, Canada},
  pages        = {10794--10803},
  year         = {2018},
  url          = {https://proceedings.neurips.cc/paper/2018/hash/53edebc543333dfbf7c5933af792c9c4-Abstract.html},
  bibsource    = {dblp computer science bibliography, https://dblp.org}
}

@inproceedings{cheng2020,
  author       = {Zhengxue Cheng and
                  Heming Sun and
                  Masaru Takeuchi and
                  Jiro Katto},
  title        = {Learned Image Compression With Discretized Gaussian Mixture Likelihoods
                  and Attention Modules},
  booktitle    = {2020 {IEEE/CVF} Conference on Computer Vision and Pattern Recognition,
                  {CVPR} 2020, Seattle, WA, USA, June 13-19, 2020},
  pages        = {7936--7945},
  publisher    = {Computer Vision Foundation / {IEEE}},
  year         = {2020},
  url          = {https://openaccess.thecvf.com/content\_CVPR\_2020/html/Cheng\_Learned\_Image\_Compression\_With\_Discretized\_Gaussian\_Mixture\_Likelihoods\_and\_Attention\_CVPR\_2020\_paper.html},
  doi          = {10.1109/CVPR42600.2020.00796},
  bibsource    = {dblp computer science bibliography, https://dblp.org}
}

@inproceedings{resnet_2016,
  author       = {Kaiming He and
                  Xiangyu Zhang and
                  Shaoqing Ren and
                  Jian Sun},
  title        = {Deep Residual Learning for Image Recognition},
  booktitle    = {2016 {IEEE} Conference on Computer Vision and Pattern Recognition,
                  {CVPR} 2016, Las Vegas, NV, USA, June 27-30, 2016},
  pages        = {770--778},
  publisher    = {{IEEE} Computer Society},
  year         = {2016},
  url          = {https://doi.org/10.1109/CVPR.2016.90},
  doi          = {10.1109/CVPR.2016.90},
  bibsource    = {dblp computer science bibliography, https://dblp.org}
}

@article{shannon_1948,
  author  = {Shannon, Claude E.},
  title   = {A Mathematical Theory of Communication},
  journal = {Bell System Technical Journal},
  year    = {1948},
  volume  = {27},
  number  = {3},
  pages   = {379--423},
  note    = {Part II: 27(4): 623--656}
}

@inproceedings{adigitalwatermark_1994,
  author       = {Ron G. van Schyndel and
                  Andrew Z. Tirkel and
                  Charles F. Osborne},
  title        = {A Digital Watermark},
  booktitle    = {Proceedings 1994 International Conference on Image Processing, Austin,
                  Texas, USA, November 13-16, 1994},
  pages        = {86--90},
  publisher    = {{IEEE} Computer Society},
  year         = {1994},
  url          = {https://doi.org/10.1109/ICIP.1994.413536},
  doi          = {10.1109/ICIP.1994.413536},
  bibsource    = {dblp computer science bibliography, https://dblp.org}
}

@inproceedings{featurespaceperturbation_2019,
  author       = {Nathan Inkawhich and
                  Wei Wen and
                  Hai (Helen) Li and
                  Yiran Chen},
  title        = {Feature Space Perturbations Yield More Transferable Adversarial Examples},
  booktitle    = {{IEEE} Conference on Computer Vision and Pattern Recognition, {CVPR}
                  2019, Long Beach, CA, USA, June 16-20, 2019},
  pages        = {7066--7074},
  publisher    = {Computer Vision Foundation / {IEEE}},
  year         = {2019},
  url          = {http://openaccess.thecvf.com/content\_CVPR\_2019/html/Inkawhich\_Feature\_Space\_Perturbations\_Yield\_More\_Transferable\_Adversarial\_Examples\_CVPR\_2019\_paper.html},
  doi          = {10.1109/CVPR.2019.00723},
  bibsource    = {dblp computer science bibliography, https://dblp.org}
}

@article{aucroc_1997,
  author       = {Andrew P. Bradley},
  title        = {The use of the area under the {ROC} curve in the evaluation of machine
                  learning algorithms},
  journal      = {Pattern Recognit.},
  volume       = {30},
  number       = {7},
  pages        = {1145--1159},
  year         = {1997},
  url          = {https://doi.org/10.1016/S0031-3203(96)00142-2},
  doi          = {10.1016/S0031-3203(96)00142-2},
  bibsource    = {dblp computer science bibliography, https://dblp.org}
}
